\pdfoutput=1
\documentclass{article}

\usepackage{microtype}
\usepackage{graphicx}
\usepackage{booktabs}
\usepackage{array}
\usepackage{multirow}
\usepackage{amsmath,amssymb,amsthm,mathtools}
\usepackage{xspace}
\usepackage[hypertexnames=false]{hyperref}
\usepackage{algorithm}
\usepackage{algorithmic}
\usepackage{stfloats}
\usepackage{capt-of}
\usepackage{enumitem}
\usepackage{placeins}
\usepackage{tikz}
\usetikzlibrary{arrows.meta,calc,backgrounds,fit}
\usepackage{pgfplots}
\pgfplotsset{compat=1.18}
\definecolor{cerboblue}{HTML}{0072B2}
\definecolor{rabitorange}{HTML}{E69F00}

\usepackage[accepted]{mlsys2025}
\makeatletter
\renewcommand{\mlsys@appearing}{Preprint. \today.}
\makeatother
\AtBeginDocument{\setlength{\footskip}{25pt}}
\fancypagestyle{firstpage}{\fancyhf{}\renewcommand{\headrulewidth}{0pt}\fancyfoot[C]{\thepage}}
\hypersetup{pdfsubject={},pdfauthor={Farhan Shahriyar Hossain, Taufikur Rahman Fuad, Md. Zarif Bin Hasnat, Md Abrar Jahin, Md Rizwan Parvez}}
\usepackage{xurl}  % lets long URLs in the references break at any character
\providecommand{\doi}[1]{\href{https://doi.org/#1}{doi:\nolinkurl{#1}}}

\newtheorem{theorem}{Theorem}
\newtheorem{proposition}[theorem]{Proposition}
\newtheorem{lemma}[theorem]{Lemma}
\newtheorem{corollary}[theorem]{Corollary}
\theoremstyle{remark}

\newcommand{\R}{\mathbb{R}}
\newcommand{\ours}{CERBO\xspace}
\newcommand{\acr}[1]{\underline{\textbf{#1}}}
\xspaceaddexceptions{\,}
\newcommand{\auc}{\mathrm{AUC}_1}

\mlsystitlerunning{CERBO: Certified Recourse at a Requested Rate}

\begin{document}
% columns end where their text ends instead of stretching the space between paragraphs
\raggedbottom

\twocolumn[
\mlsystitle{CERBO: Certified Recourse at a Requested Rate for Gradient-Boosted Trees}

\begin{mlsysauthorlist}
\mlsysauthor{Farhan Shahriyar Hossain}{iut}
\mlsysauthor{Taufikur Rahman Fuad}{iut}
\mlsysauthor{Md. Zarif Bin Hasnat}{iut}
\mlsysauthor{Md Abrar Jahin}{usc}
\mlsysauthor{Md Rizwan Parvez}{qcri}
\end{mlsysauthorlist}
\mlsysaffiliation{iut}{Islamic University of Technology}
\mlsysaffiliation{usc}{University of Southern California}
\mlsysaffiliation{qcri}{Qatar Computing Research Institute (QCRI)}
\mlsyscorrespondingauthor{Farhan Shahriyar Hossain}{farhanhossain@iut-dhaka.edu}

\mlsyskeywords{Algorithmic Recourse, Counterfactual Explanations, Gradient-Boosted Decision Trees, Tree Ensembles, Certification, Calibration}

\vskip 0.3in

\begin{abstract}
Recourse-aware learning trains models so that people who receive an adverse
decision can reverse it with an affordable change. For gradient-boosted decision
trees (GBDTs), the existing method, RABIT, exposes a single recourse weight whose
effect on the recourse rate is abrupt and data-dependent, and its training bound
cannot certify recourse for any rejected individual. We present \acr{Ce}rtified \acr{R}ecourse \acr{Bo}osting
(\ours), which takes a requested recourse rate and returns a GBDT with
certified actions: it tracks exact additive effort along tree paths, trains
against committed boxes (a family of bounds that contains RABIT's and certifies a
within-budget action for each training sample) and adjusts the weight after every
tree until the certified rate meets the request. A calibration step guarantees
the rate on new individuals, and the certifier runs on any tree ensemble,
including XGBoost and LightGBM, in milliseconds per person without a solver.
In RABIT's own experimental protocol, at equal accuracy, \ours raises the
recourse ratio from $0.90$ to $0.99$ on FICO and from $0.64$ to $0.77$ on Bail.
Under additive costs on six data sets, it raises the area under the
accuracy--recourse frontier in all 12 settings, and in simulated deployments it guarantees
recourse to 70\% of rejected individuals in 11 of 12 settings, against 2 for
RABIT.
\end{abstract}
]
\thispagestyle{firstpage}

% affiliations, correspondence and notice in the first-page footnote, as the template places them
{\let\thefootnote\relax\footnotetext{\hspace*{-1.8em}\parindent=0pt\raggedright
\textsuperscript{1}Islamic University of Technology
\textsuperscript{2}University of Southern California
\textsuperscript{3}Qatar Computing Research Institute (QCRI).
Correspondence to: Farhan Shahriyar Hossain
\textless{}farhanhossain@iut-dhaka.edu\textgreater{}.\\
Preprint. \today.}}

\section{Introduction}
Algorithmic recourse asks which feasible, affordable changes would reverse an
unfavorable prediction \citep{wachter2018,ustun2019,karimi2021}. Post-hoc
methods search for such changes in a fixed model
\citep{karimi2020,mothilal2020,parmentier2021}, but a model trained only for
accuracy may admit none \citep{ross2021,kanamori2024}. Recourse-aware learning
therefore changes training itself. For gradient-boosted decision trees (GBDTs),
the dominant model family on tabular data \citep{grinsztajn2022}, RABIT
\citep{kanamori2025rabit} adds a recourse loss with weight $\gamma$ to the
second-order boosting objective.

\begin{figure*}[t]
\centering
\includegraphics[width=\linewidth]{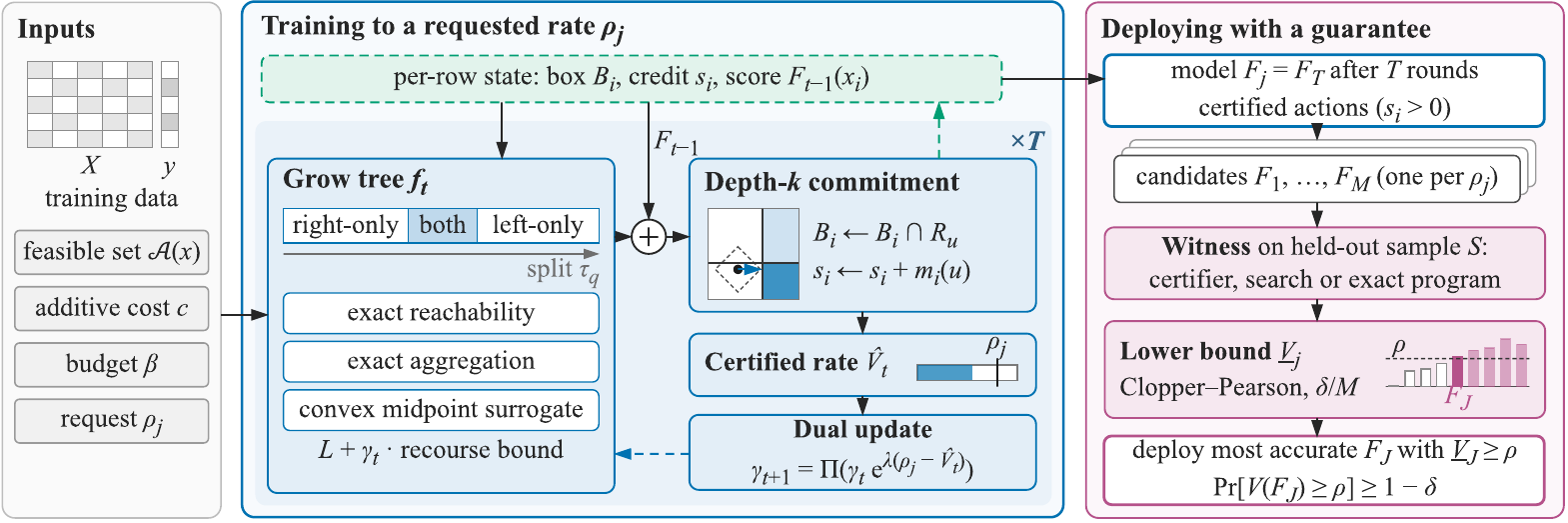}
\caption{\ours trains one model per requested rate and deploys one with a guaranteed
rate. Each training row keeps a box $B_i$, a credit $s_i$ and
its score. In each of $T$ boosting rounds, \ours grows tree $f_t$ on the loss $L$
plus $\gamma_t$ times a recourse bound, using exact reachability, exact aggregation
over the right-only, both and left-only reach states of a split, and a convex
surrogate. Each row then commits to a reachable node $u$, a positive credit
certifies an action, and the certified rate $\hat V_t$ of rejected rows sets
$\gamma_{t+1}$. Deployment keeps the most accurate candidate whose
Clopper--Pearson lower bound $\underline V_j$ on a held-out sample reaches $\rho$;
its rate on new individuals is then at least $\rho$ with probability at least
$1-\delta$. The bars show the candidates' lower bounds by requested rate for one
calibration draw on Adult at $\beta=0.2$. Sections~\ref{sec:exact}
to~\ref{sec:deploy} describe the four steps.}
\label{fig:overview}
\end{figure*}

RABIT leaves three practical gaps. The first is \emph{how much} recourse a
weight buys. The share of rejected people with recourse switches abruptly over a
narrow range of $\gamma$, at a location that differs across data sets, and we
prove that weighted training returns only models on the lower convex hull of
the accuracy--recourse trade-off. A practitioner therefore cannot ask for
``recourse for 70\% of rejected applicants'' and get it.

The second gap is \emph{who} has recourse. RABIT's training bound treats every
rejected sample as having none, so the learner cannot tell whose recourse it
creates, and nothing guarantees a recourse rate on new people. Both questions
matter in deployment: a US lender must tell a rejected applicant the specific
reasons for the decision \citep{regb}, and the EU AI Act gives people a right to
an explanation of decisions made with high-risk systems such as credit scoring
\citep{euaiact}. An operator who also promises recourse must know for what share
of rejected people the promise holds.

The third gap is the effort model. RABIT's split search assumes a max-type cost,
under which an action is affordable exactly when every per-feature change is,
yet effort is often additive: raising income and reducing debt costs more than
either alone. Its authors leave additive budgets to future work
\citep[Section~6]{kanamori2025rabit}.

We present \acr{Ce}rtified \acr{R}ecourse \acr{Bo}osting (\ours). A practitioner states a
recourse rate and an effort budget, and \ours returns a GBDT with certified
actions, as Figure~\ref{fig:overview} illustrates. Our contributions
are:
\begin{enumerate}[leftmargin=*,itemsep=1pt,topsep=2pt]
\item \emph{Exact training under additive costs.} Section~\ref{sec:exact} tracks
each sample's remaining budget along tree paths and aggregates split statistics
exactly in $O(r\log K+K)$ time for $r$ samples and $K$ thresholds, and
replaces RABIT's indefinite surrogate with a convex one.
\item \emph{Certified recourse.} Committed boxes (Section~\ref{sec:cb}) are a
family of training bounds that contains RABIT's as its weakest member and
certifies a within-budget action for each sample; the same certifier audits any
tree ensemble, including XGBoost and LightGBM, in milliseconds per person.
\item \emph{Recourse at a requested rate.} A dual update drives the certified
training rate to the request, and a calibration step guarantees the rate on new
individuals with probability $1-\delta$ (Sections~\ref{sec:target}
and~\ref{sec:deploy}).
\item \emph{Experiments in two settings.} In RABIT's own experimental
protocol, \ours matches RABIT's accuracy, raises the recourse ratio on every data
set and finds cheaper actions; under additive costs it raises the area under the
accuracy--recourse frontier in all 12 settings (Section~\ref{sec:exp}).
\end{enumerate}

% ---------------------------------------------------------------------------
\section{Background}
\label{sec:background}
\textbf{Setting. }
We consider binary classification with favorable class $+1$ and ensembles
$F_T=\sum_{t\le T}f_t$ of $T$ regression trees. Feasible actions factor by
coordinate, $\mathcal A(x)=\prod_d\mathcal A_d(x_d)$, which covers immutable,
increase-only and decrease-only features. Costs are additive and
coordinate-separable, $c(a\mid x)=\sum_d\phi_d(a_d;x_d)$, with each $\phi_d$
non-decreasing in the distance moved. Our experiments use the percentile shift
$|Q_d(x_d+a_d)-Q_d(x_d)|$, where $Q_d$ is the cumulative distribution function
of feature $d$. RABIT's split search instead assumes the max-type
cost $\max_d\phi_d(a_d;x_d)$; its experiments use the max percentile shift.

For a budget $\beta$, let $\mathcal A_\beta(x)=\{a\in\mathcal A(x):
c(a\mid x)\le\beta\}$. The recourse loss of $x$ is
$\ell_\beta(x\mid F)=\min_{a\in\mathcal A_\beta(x)}\ell(F(x+a))$, where
$\ell(s)=\ell(+1,s)$ is a non-increasing margin loss (the logistic loss in our
experiments), and $x$ has recourse if $F(x+a)>0$ for some $a\in\mathcal A_\beta(x)$.
We measure the share of rejected people with recourse by a verified search
(validity) and by certificates (certified rate).

\textbf{RABIT. }
RABIT grows tree $t$ greedily to minimize the training loss plus $\gamma$ times
an upper bound on $\sum_i\ell_\beta(x_i)$ over rejected samples. The
bound replaces the earlier trees by
$\xi_t(x)=\sum_{t'<t}\min_{a\in\mathcal A_\beta(x)}f_{t'}(x+a)$, the sum of the
worst leaf values reachable in each earlier tree. A sample that can reach only
one child of a split contributes the loss gradient to that child. A sample that
can reach both contributes through a soft-min of the two child losses. A
second-order expansion gives closed-form leaf values and split gains. After
training, an optional leaf refinement refits leaf values with a logistic
regression that also rewards found actions.

% ---------------------------------------------------------------------------
\section{\ours}
\label{sec:method}
\ours keeps RABIT's boosting loop, changes the bound each tree is trained against,
and adds a rate controller and a deployment step, as Figure~\ref{fig:overview}
shows. Appendix~\ref{app:proofs} proves every result.

\subsection{Exact Additive Training}
\label{sec:exact}
\textbf{Exact reachability. }
For a box $B=\prod_dI_d$, let $\kappa_d(x;I_d)$ be the minimum one-dimensional
cost of moving $x_d$ into the interval $I_d$, and $C(x,B)=\sum_d\kappa_d(x;I_d)$.

\begin{proposition}[Box cost]\label{prop:box}
$\min\{c(a\mid x):a\in\mathcal A(x),\,x+a\in B\}=C(x,B)$.
\end{proposition}

\begin{proposition}[Independent caps]\label{prop:caps}
For $D\ge2$ features, no product set $\prod_d[0,b_d]$ equals the additive budget
set $\{c\in\R_+^D:\sum_dc_d\le\beta\}$.
\end{proposition}

A product set cannot cut the diagonal face of the budget simplex, so
per-feature caps either admit unaffordable actions or exclude affordable ones.

When a split on feature $d$ intersects the current interval with a half-line,
the $d$-term becomes the maximum of the old term and the crossing cost, by
Lemma~\ref{lem:max} in Appendix~\ref{app:proofs}. The path cost updates
in constant time per sample, and a sample reaches a child exactly when its
updated cost is at most $\beta$.

\textbf{Exact aggregation. }
For threshold $\tau_q$ on feature $d$, sample $i$ reaches the left child if
$x_{id}\le\tau_q$ or its crossing cost is within its remaining budget $b_i$, and
symmetrically for the right child. Under additive costs, $b_i$ differs across
samples.

\begin{proposition}[Monotone states]\label{prop:mono}
As $q$ increases, sample $i$ is right-only, then reaches both children, then is
left-only. It never reaches neither.
\end{proposition}

\begin{corollary}[Exact aggregation cost]\label{cor:agg}
The gradient and Hessian sums of the left-only, both and right-only sets for
all $K$ thresholds of a feature can be computed exactly in $O(r\log K+K)$ time,
where $r$ is the number of samples reaching the node.
\end{corollary}

Two binary searches locate the two transitions of each sample, and difference
arrays accumulate their contributions. A scan over samples sorted by feature
value, as used for constant caps, assumes all samples share one budget.
Under per-sample budgets the transitions are not ordered by feature value;
Appendix~\ref{app:impl} lists the differences from RABIT's code.

\textbf{A convex surrogate. }
For a sample that reaches both children, RABIT bounds
$\min(\ell(s+w_L),\ell(s+w_R))$, where $s$ bounds the earlier trees' score and
$w_L,w_R$ are the two leaf values, by a LogSumExp soft-min with temperature
$\nu$ ($\nu=1$ in the released code). At $w=0$ its second-order model, as
implemented, has diagonal $\ell''/(2\nu)$ and off-diagonal coupling
$\ell'^2/(4\nu)$.

For the logistic loss this matrix is indefinite exactly when
$\sigma(s)<1/3$, with $\sigma$ the logistic function, for every $\nu$, as
Lemma~\ref{lem:indef} in Appendix~\ref{app:proofs} shows. This happens once the
worst-case score is clearly negative, the regime recourse training targets. The two-leaf Newton system can then become singular and the leaf
values diverge. At $\gamma=0.2$ the released RABIT diverges in our runs: 13 training runs on German fail, and
11 models trained for the certification experiments (Credit, German, HMEQ) have
non-finite leaf values.

\begin{lemma}[Midpoint bound]\label{lem:mid}
For any non-increasing $\ell$,
$\min(\ell(s+w_L),\ell(s+w_R))\le\ell\big(s+\tfrac{w_L+w_R}{2}\big)$. The right-hand
side is convex in $(w_L,w_R)$ for convex $\ell$; at $0$ its Hessian is
$\tfrac{\ell''}{4}\begin{psmallmatrix}1&1\\1&1\end{psmallmatrix}\succeq0$.
\end{lemma}

With the midpoint bound the objective remains an upper bound on the recourse
loss. The two-leaf system is positive definite, because the loss Hessians and
the ridge term $\alpha>0$ add to its diagonal, so leaf values are well
defined.

\subsection{Committed Boxes and Certified Recourse}
\label{sec:cb}
RABIT's bound $\xi$ takes the worst reachable leaf of each earlier tree
independently. The worst leaves of different trees are rarely reachable by one
action, so the bound is loose, and it gives nearly every sample the same
recourse gradient. We instead keep, for each training sample, a box $B_i$, the
intersection of the tree regions it has committed to, and a credit $s_i$, with
the invariant
\begin{equation}\label{eq:inv}
F_{t-1}(x_i+a)\ge s_i\ \ \text{for all }a\in\mathcal A_\beta(x_i)\text{ with }x_i+a\in B_i .
\end{equation}

\textbf{Depth-$k$ commitment. }
After tree $t$ is grown, sample $i$ commits to a node $u$ at depth $k$ (or a
shallower leaf) whose region $R_u$ is reachable from $B_i$ within budget. It chooses the node with the largest $m_i(u)$, the minimum value over the leaves
below $u$ that remain reachable. Then $B_i\leftarrow B_i\cap R_u$ and
$s_i\leftarrow s_i+m_i(u)$. When growing tree $t$, the recourse term uses
$s_i$ as the score bound $s$ of the surrogate and the reachability of $B_i$ in
place of $\xi_t$.

\begin{theorem}[Commitment invariant]\label{thm:valid}
Depth-$k$ commitment preserves \eqref{eq:inv} for every $k$. Hence, for every new
tree $h$,
\[
\ell_\beta(x_i\mid F_{t-1}+h)\le\min_{\substack{a\in\mathcal A_\beta(x_i)\\x_i+a\in B_i}}\ell\big(s_i+h(x_i+a)\big).
\]
For $k=0$ the box never shrinks and $s_i=\xi_t(x_i)$, RABIT's bound; for $k$ at
least the tree depth, $F_{t-1}=s_i$ exactly on $B_i$.
\end{theorem}

\begin{theorem}[Training certificate]\label{thm:cert}
Let $a_i$ be the projection of $x_i$ onto $B_i$. Then $a_i\in\mathcal A(x_i)$,
$c(a_i\mid x_i)\le\beta$ and $F_T(x_i+a_i)\ge s_i$. Thus $s_i>0$ certifies a valid
within-budget action, without solving an optimization problem.
\end{theorem}

By Proposition~\ref{prop:kmono} in Appendix~\ref{app:proofs}, credit is
monotone in $k$ for one step, but deeper commitment leaves smaller boxes, in which
later trees reach fewer leaves. We use $k=1$.

\textbf{Test-time certification. }
Theorem~\ref{thm:cert} does not depend on how the trees were trained.

\begin{theorem}[Test-time certificate]\label{thm:test}
For any tree ensemble $F_T$ and any $x$, run depth-$k$ commitment through the
trees of $F_T$ starting from $B=\R^D$ and $s=0$. The resulting box $B$ and credit
$s$ satisfy $F_T(x+a)\ge s$ for all $a\in\mathcal A_\beta(x)$ with $x+a\in B$,
and the projection of $x$ onto $B$ is feasible with cost at most $\beta$. If
$s>0$, this action is valid recourse. The procedure takes $O(T\,2^k D)$ time.
\end{theorem}

\textbf{Three sound witnesses. }
The certifier is sound but incomplete: $s\le0$ does not prove that no recourse
exists. A verified action is feasible, costs at most $\beta$ and is approved by
the model; a sound \emph{witness} returns only verified actions. We use three: RABIT's heuristic search over leaf regions, the
certifier above, and an exact mixed-integer program, whose running times
Section~\ref{sec:libs} reports. Section~\ref{sec:ablate} shows that search finds
far more actions than the certifier and nearly all that exist, so the certifier
does not replace search at test time. It is the bound \ours
trains with, which makes the certified rate visible during training.

\begin{figure}[t]
\centering
\includegraphics[width=\linewidth]{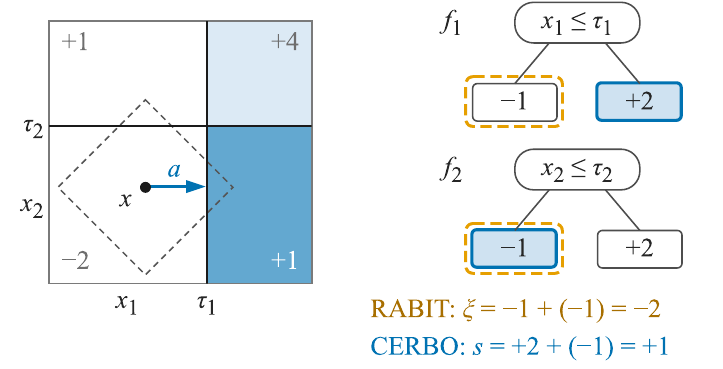}
\caption{A committed box certifies an action that RABIT's bound misses. Two stumps
$f_1$ and $f_2$ (left branch: split condition holds) under an additive budget;
cells show $F=f_1+f_2$, and the dashed region holds the points $x+a$ with
$c(a\mid x)\le\beta$. The point $x$ can cross either split within the budget, but
not both.
RABIT's bound adds the worst reachable leaf of each tree (orange) and certifies
nothing. With $k=1$, tree~1 commits $x$ to $x_1>\tau_1$ (light blue) and tree~2 to
its only reachable leaf, $x_2\le\tau_2$, giving the dark-blue box; committed
leaves are blue. The credit $s=+1$ certifies the projection $a$ of $x$ onto this box, with
$F(x+a)=1$.}
\label{fig:commit}
\end{figure}

\textbf{Relation to RABIT's bound. }
Committed boxes contain RABIT's bound as their weakest member, and the gap
between the two can be as large as the ensemble's range, as
Figure~\ref{fig:commit} and Proposition~\ref{prop:gap} show. Let $\xi=\xi_{T+1}$ be
RABIT's bound over all $T$ trees of a fixed ensemble, with exact reachability,
and $s_k(x)$ the credit of
Theorem~\ref{thm:test}.

\begin{theorem}[Sandwich]\label{thm:sandwich}
For every tree ensemble $F$, budget $\beta$, input $x$ and depth $k\ge0$,
\begin{gather*}
\xi(x)\le\min_{a\in\mathcal A_\beta(x)}F(x+a)\le F(x),\\
\xi(x)=s_0(x)\le s_k(x)\le\max_{a\in\mathcal A_\beta(x)}F(x+a).
\end{gather*}
\end{theorem}

\begin{corollary}[RABIT's bound cannot certify a rejected individual]\label{cor:nocert}
If $F(x)\le0$ then $\xi(x)\le0$; hence RABIT's bound certifies no rejected
individual.
\end{corollary}

RABIT's bound lower-bounds the score of the \emph{worst} affordable action, and
``do nothing'' is always affordable. Certificates require commitment ($k\ge1$),
which bounds the score of a \emph{chosen} action.

\begin{proposition}[Worst-case gap]\label{prop:gap}
For every $T\ge1$ and $w>0$ there is a one-feature ensemble of $T$ stumps and a
rejected $x$ with $\xi(x)=F(x)=-Tw$ and $s_1(x)=\max_{a\in\mathcal A_\beta(x)}F(x+a)=Tw$.
\end{proposition}

\subsection{Training to a Requested Rate}
\label{sec:target}
\textbf{Why a fixed weight cannot express a rate. }
A practitioner's requirement is a constraint: the most accurate model whose
recourse rate is at least $\rho$,
\begin{equation}\label{eq:req}
\mathrm{P}(\rho):\quad\min_{F\in\mathcal F}L(F)\quad\text{s.t.}\quad V(F)\ge\rho ,
\end{equation}
where $L$ is the training loss, $V$ the recourse rate and $\mathcal F$ the
models the learner can produce. Weighted training instead returns
$F_\gamma\in\arg\min_{F\in\mathcal F}L(F)+\gamma\,\Omega(F)$ for a recourse loss $\Omega$.
Let $\mathcal P=\{(L(F),\Omega(F)):F\in\mathcal F\}$.

\begin{proposition}[Weights reach only the convex hull]\label{prop:weights}
(i) For every $\gamma\ge0$, $(L(F_\gamma),\Omega(F_\gamma))$ lies on the lower convex hull
of $\mathcal P$; models above the hull are never returned for any $\gamma$.
(ii) $\Omega(F_\gamma)$ is non-increasing and $L(F_\gamma)$ non-decreasing in $\gamma$,
and both change only where the minimizing hull vertex changes, so the returned
rate moves in jumps between hull vertices.
(iii) There are classes of three models in which, for every
$\rho\in(0,\tfrac12]$, the solution of \eqref{eq:req} is never returned: every
$\gamma$ yields either a model with rate $0<\rho$ or one that is strictly less
accurate than necessary, and the rate jumps from $0$ to $1$ at a single $\gamma$.
\end{proposition}

Parts (i) and (ii) restate known limits of weighted-sum scalarization
\citep{das1997}. Tree
ensembles form a discrete, non-convex class, so such gaps are to be expected. A randomized mixture of two hull models could fill a gap, but randomized
credit or bail decisions are unacceptable in practice and admit no per-person
certificate.

Our experiments match this picture: across 18 weights, the released RABIT's
validity stays below $0.31$ on Adult and Bail at $\beta=0.1$, and
Figure~\ref{fig:weights} in the appendix shows how abruptly it changes in
$\gamma$.

\textbf{Closed-loop targeting. }
\ours does not search over a weight. Certificates are a by-product of training,
so the certified rate among currently rejected training samples,
$\hat V_t=\Pr_i[s_i>0\mid F_t(x_i)\le0]$, is known exactly after every tree. For a
target rate $\rho$ we update
\begin{equation}\label{eq:dual}
\gamma_{t+1}=\Pi_{[\gamma_{\min},\gamma_{\max}]}\big(\gamma_t\exp(\lambda(\rho-\hat V_t))\big),
\end{equation}
a multiplicative (mirror) ascent step with step size $\lambda$ on the dual of
``minimize the training loss subject to certified rate at least $\rho$''.
Dual updates of this kind are standard in constrained learning
\citep{cotter2019,chamon2023}; the new element is that certificates make the
constrained quantity observable after every tree. Algorithm~\ref{alg:cerbo} in
Appendix~\ref{app:impl} lists the training loop.

\begin{proposition}[Rate tracking]\label{prop:track}
If the projection in \eqref{eq:dual} is inactive for $t=1,\dots,T$, then
\[
\Big|\frac1T\sum_{t=1}^{T}\hat V_t-\rho\Big|\le\frac{\log(\gamma_{\max}/\gamma_{\min})}{\lambda T}.
\]
\end{proposition}

\subsection{Deploying with a Guaranteed Rate}
\label{sec:deploy}
Fix any sound witness $w_F(x)\in\{0,1\}$, which returns $1$ only with a
verified action (the certifier, RABIT's search or an exact
program; it may use fixed internal randomness). For a model $F$, let
$V(F)=\Pr[w_F(x)=1\mid F(x)\le0]$ be its witnessed recourse rate on new
individuals from the data distribution $P$.

Train candidate models
$F_1,\dots,F_M$ (for example, one per requested rate $\rho_j$;
Figure~\ref{fig:overview}) and draw a
calibration sample $S$ of $N$ individuals from $P$, independent of training. For
model $j$, let $n_j$ be the number of individuals in $S$ that $F_j$ rejects
and $z_j$ the number of them with a witness. Let $\underline V_j$ be a lower
confidence bound at level $\delta/M$ on the proportion that $z_j/n_j$ estimates, such as
Hoeffding's or the exact Clopper--Pearson bound \citep{clopper1934}, which we use.

\begin{theorem}[Recourse-rate control]\label{thm:control}
Let $J$ be any selection rule, possibly depending on $S$ and on other data, that
selects only models with $\underline V_J\ge\rho$ (for example, the most accurate such
model). Then, with probability at least $1-\delta$ over $S$, $V(F_J)\ge\rho$,
and every witnessed individual receives a feasible action of cost at most
$\beta$ that $F_J$ approves.
\end{theorem}

Like conformal prediction
\citep{vovk2005,angelopoulos2021gentle} and risk control by multiple testing
\citep{angelopoulos2021ltt}, the recipe is distribution-free and holds in finite
samples when calibration and deployment individuals come from the same
distribution. \ours adds candidates at requested rates, rather than at the
scattered rates that, by Proposition~\ref{prop:weights}, a weight produces. RABIT's
probably approximately correct (PAC) bound \citep[Prop.~3]{kanamori2025rabit}
instead bounds the chance that one model admits no affordable action, through a
training surrogate and a capacity term.

\subsection{Implementation}
\label{sec:impl}
\ours extends RABIT's public Python implementation and reuses its tree, binning
and action code (Algorithm~\ref{alg:cerbo} in Appendix~\ref{app:impl}). Each
training row carries its box bounds, per-feature path costs and credit, so the
state grows linearly with the data. Numba \citep{lam2015numba} compiles the split
search, which performs the exact aggregation of Corollary~\ref{cor:agg}. After each tree, one vectorized step commits all rows: a
depth-first walk carries every row's reachability and path cost to each node,
and a bottom-up pass takes the minimum reachable leaf value below each node.
The test-time certifier runs the same step from an unconstrained box on any
ensemble stored as RABIT's tree arrays, to which converters map XGBoost and
LightGBM models.

% ---------------------------------------------------------------------------
\section{Experiments}
\label{sec:exp}
\begin{table*}[t]
\centering
\caption{\ours under RABIT's experimental protocol (max percentile-shift cost with kernel
density estimate (KDE) percentiles, $\beta=0.2$). Values are test means over ten
75/25 splits, with the standard deviation as a subscript for accuracy and
recourse ratio. OAF boosts on actionable features only. RABIT appears at
its published $\gamma$ and with $\gamma$ tuned by the same validation rule as
\ours's $\rho$. The best value per data set is in bold for the recourse
ratio and the three action metrics; accuracy, ROC-AUC and F1 have no bold,
because Vanilla, trained for accuracy alone, is their reference. On Bail the rule selects $\gamma=0.002$, so
the two RABIT rows coincide. \ours raises the recourse ratio on every data set
at equal accuracy. $^\dagger$Cost-model artifact, not counted
(Section~\ref{sec:setup}).}
\label{tab:rabitbench}
\small
\setlength{\tabcolsep}{4pt}
\begin{tabular}{llccccccc}
\toprule
Data & Method & Accuracy $\uparrow$ & Recourse ratio $\uparrow$ & Cost $\downarrow$ & Sparsity $\downarrow$ & Plausibility $\downarrow$ & ROC-AUC $\uparrow$ & F1 $\uparrow$\\
\midrule
\multirow{5}{*}{FICO}
 & Vanilla & $0.734_{\pm.006}$ & $0.542_{\pm.008}$ & 0.372 & 2.24 & \textbf{0.440} & 0.801 & 0.717 \\
 & OAF & $0.722_{\pm.009}$ & $0.628_{\pm.010}$ & 0.316 & 2.77 & 0.443 & 0.786 & 0.704 \\
 & RABIT ($\gamma=0.002$) & $0.728_{\pm.009}$ & $0.832_{\pm.014}$ & 0.167 & 1.05 & 0.475 & 0.796 & 0.711 \\
 & RABIT (tuned $\gamma$) & $0.727_{\pm.008}$ & $0.900_{\pm.014}$ & 0.113 & \textbf{1.02} & 0.475 & 0.790 & 0.708 \\
 & \ours & $0.728_{\pm.010}$ & $\mathbf{0.992}_{\pm.002}$ & \textbf{0.052} & 1.05 & 0.465 & 0.794 & 0.717 \\
\midrule
\multirow{5}{*}{COMPAS}
 & Vanilla & $0.684_{\pm.008}$ & $0.826_{\pm.009}$ & 0.232 & 1.51 & \textbf{0.436} & 0.736 & 0.727 \\
 & OAF & $0.685_{\pm.010}$ & $0.854_{\pm.019}$ & 0.179 & 1.55 & 0.445 & 0.734 & 0.726 \\
 & RABIT ($\gamma=0.002$) & $0.683_{\pm.009}$ & $0.932_{\pm.021}$ & 0.108 & 1.20 & 0.471 & 0.734 & 0.726 \\
 & RABIT (tuned $\gamma$) & $0.681_{\pm.008}$ & $0.966_{\pm.009}$ & 0.087 & 1.09 & 0.486 & 0.729 & 0.726 \\
 & \ours & $0.680_{\pm.009}$ & $\mathbf{0.990}_{\pm.007}$ & \textbf{0.071} & \textbf{1.08} & 0.491 & 0.731 & 0.734 \\
\midrule
\multirow{5}{*}{Adult}
 & Vanilla & $0.853_{\pm.002}$ & $0.273_{\pm.010}$ & 0.353 & 1.50 & 0.476 & 0.909 & 0.632 \\
 & OAF & $0.833_{\pm.004}$ & $0.307_{\pm.007}$ & 0.309 & 1.51 & 0.462 & 0.858 & 0.554 \\
 & RABIT ($\gamma=0.002$) & $0.852_{\pm.002}$ & $0.437_{\pm.031}$ & 0.281 & 1.42 & 0.478 & 0.906 & 0.633 \\
 & RABIT (tuned $\gamma$) & $0.848_{\pm.003}$ & $0.538_{\pm.023}$ & 0.240 & \textbf{1.30} & 0.479 & 0.898 & 0.627 \\
 & \ours & $0.850_{\pm.002}$ & $\mathbf{0.999}_{\pm.001}^{\,\dagger}$ & \textbf{0.083} & 1.35 & \textbf{0.439} & 0.904 & 0.633 \\
\midrule
\multirow{5}{*}{Bail}
 & Vanilla & $0.707_{\pm.006}$ & $0.366_{\pm.020}$ & 0.456 & 1.44 & 0.507 & 0.739 & 0.499 \\
 & OAF & $0.678_{\pm.006}$ & $0.410_{\pm.021}$ & 0.371 & 1.60 & \textbf{0.502} & 0.706 & 0.453 \\
 & RABIT ($\gamma=0.002$) & $0.702_{\pm.005}$ & $0.637_{\pm.025}$ & 0.225 & 1.13 & 0.517 & 0.729 & 0.474 \\
 & RABIT (tuned $\gamma$) & $0.702_{\pm.005}$ & $0.637_{\pm.025}$ & 0.225 & 1.13 & 0.517 & 0.729 & 0.474 \\
 & \ours & $0.702_{\pm.007}$ & $\mathbf{0.769}_{\pm.027}$ & \textbf{0.167} & \textbf{1.07} & 0.513 & 0.731 & 0.511 \\
\bottomrule
\end{tabular}
\end{table*}

Sections~\ref{sec:rabitbench} to~\ref{sec:ablate} test five claims in turn:
\ours outperforms RABIT under RABIT's own protocol and along the whole trade-off
under additive effort, requested rates are met and can be guaranteed, certifying
a deployed ensemble is cheap, and each component contributes.

\subsection{Experimental Setup}
\label{sec:setup}
\textbf{Data sets. }
All seven data sets are public; we give individuals $\times$ encoded features
below, and Table~\ref{tab:data} in the appendix lists sources and licenses. The
first setting follows RABIT's experimental protocol on FICO \citep{fico2018}
($9{,}871\times23$), COMPAS \citep{angwin2016} ($7{,}214\times15$), Adult
\citep{becker1996adult} ($48{,}842\times16$) and Bail \citep{schmidt1988}
($8{,}923\times16$), with RABIT's actionability metadata and evaluation code. The
second compares whole trade-offs under additive costs on Adult, Bail, FICO and
Credit \citep{yeh2009} ($30{,}000\times13$; RABIT's data and metadata, Credit
from its code release), and on German Credit \citep{hofmann1994}
($1{,}000\times51$) and HMEQ \citep{baesens2016} ($5{,}960\times18$), two
held-out data sets added after method development. COMPAS is absent from the
second setting because its mutable one-hot group is outside our exactness
results for additive budgets.

\textbf{Baselines. }
We compare with Vanilla (boosting without recourse), OAF (boosting on actionable
features only, RABIT's second baseline) and RABIT, to the best of our knowledge
the only published method that trains GBDTs for recourse. All methods share
RABIT's booster and configuration: 100 trees of depth 8, learning rate $0.1$,
$\sqrt D$ candidate features per split, 128 bins and ridge term $\alpha=0.001$.
This implementation is within about one point of scikit-learn's tuned histogram
gradient boosting except on Bail ($0.700$ vs.\ $0.851$), and Appendix~\ref{app:rob}
shows that a stronger configuration raises Bail only to $0.733$ and widens
\ours's lead. Under additive costs, for which RABIT's algorithm is not designed,
we run the depth-scheduled budget rule of its released code, and ablations B--D
in Section~\ref{sec:ablate} also give RABIT exact accounting and aggregation.
\ours uses initial $\gamma=0.01$, $\lambda=0.5$ and $\gamma\in[10^{-3},1]$;
Table~\ref{tab:settings} in the appendix lists every setting.

\textbf{Metrics. }
Effort under additive costs is the mid-rank empirical percentile shift. RABIT's
kernel density estimate (KDE) of the percentile smooths away point masses: on
Adult, raising capital gain from $0$ to $1$ costs $0.0003$. Validity is the
share of rejected test individuals for whom RABIT's search finds a verified
action with cost at most $\beta$. Because it changes abruptly in $\gamma$, we
compare accuracy--recourse frontiers with
validity as the recourse measure. Each trained model gives one point: its
accuracy drop relative to Vanilla, in percentage points, and its validity, both
averaged over seeds. With $\operatorname{val}(d)$ the best validity at an accuracy
drop of at most $d$ points, we report the normalized area under the frontier
\begin{equation}\label{eq:auc}
\mathrm{AUC}_\Delta=\frac1\Delta\int_0^\Delta\operatorname{val}(d)\,\mathrm{d}d
\end{equation}
with $\Delta=1$, the tolerance of the validation rule in
Section~\ref{sec:rabitbench}.

RABIT gets 18 weights from $0.001$ to $0.2$,
dense around its jump (its paper uses $0.002$; some runs diverge at $0.2$), and
\ours eight targets $\rho\in\{0.2,\dots,0.9\}$. We report two frontiers: the
model frontier, the best validity of a trained model at each accuracy drop, and
the upper concave envelope, which also credits randomizing between two models.
The model frontier is stricter, since randomized decisions are unacceptable in
consequential settings. Both frontiers, and the $0.5$- and $2$-point windows of
Table~\ref{tab:window} in the appendix, lead to the same conclusions.

\textbf{Statistics. }
Each seed is a fresh random train/test split, and intervals are 95\% paired
percentile-bootstrap intervals over seeds (2{,}000 resamples), which include split and
test-sample variation. Method development used seeds 0--12 and 100--107 (0--1
on synthetic data). Every reported number comes from later seeds, which
Table~\ref{tab:runs} in the appendix lists, and no table selects a model on test
data.

\textbf{Software and hardware. }
Runs used Python 3.11, NumPy \citep{harris2020numpy}, SciPy
\citep{virtanen2020scipy}, scikit-learn 1.5
\citep{pedregosa2011}, XGBoost 2.1 \citep{chen2016}, LightGBM 4.5
\citep{ke2017lightgbm} and SCIP through PySCIPOpt 5.5 \citep{maher2016};
Appendix~\ref{app:impl} describes the hardware. The code (MIT license), all run
results and a one-command reproduction script will be released upon publication.

% ---------------------------------------------------------------------------
\subsection{Comparison under RABIT's Protocol}
\label{sec:rabitbench}
\begin{figure*}[t]
\centering
\includegraphics[width=\linewidth]{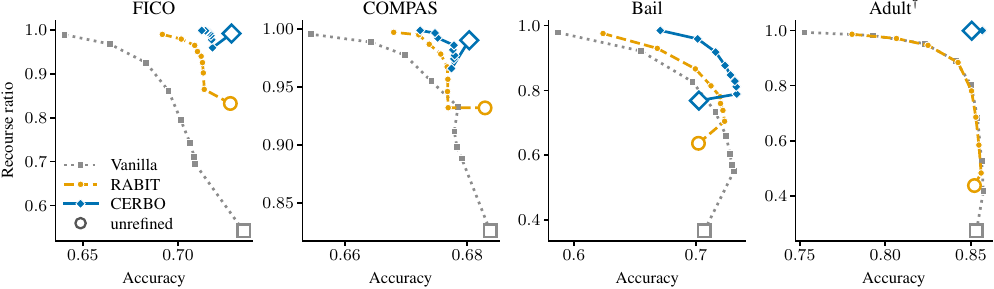}
\caption{RABIT's leaf refinement at every level $\eta$ under RABIT's protocol:
recourse ratio against test accuracy, mean over ten splits, for Vanilla, RABIT at
its published $\gamma=0.002$ and \ours. Hollow markers are the unrefined models; the other points use
$\eta=0.01$ to $1.28$. At every $\eta>0$, \ours\,+\,refinement is at least as
accurate as RABIT\,+\,refinement and has a higher recourse ratio.
$^\dagger$Cost-model artifact.}
\label{fig:rabitref}
\end{figure*}

\begin{table*}[b]
\centering
\caption{Area under the accuracy--recourse frontier (validity; $\auc$, higher
is better) under additive costs, test data, mean over seeds 200--207. OAF trains on actionable
features only, and RABIT gets 18 weights. $\Delta$ is \ours minus RABIT with a
95\% paired bootstrap interval, on the upper concave envelope (as in the $\auc$
columns) and on the model frontier, which credits only trained models. The best
$\auc$ per row is in bold. German and HMEQ (bottom) were added after method
development. \ours leads in all 12 settings.}
\label{tab:main}
\small
\begin{tabular}{lccccc}
\toprule
Setting & OAF & RABIT & \ours & $\Delta$ (envelope) & $\Delta$ (model frontier)\\
\midrule
Adult $\beta=0.1$  & 0.076 & 0.247 & \textbf{0.427} & $+0.180$ [$+0.164$, $+0.203$] & $+0.127$ [$+0.107$, $+0.153$] \\
Adult $\beta=0.2$  & 0.137 & 0.418 & \textbf{0.609} & $+0.191$ [$+0.160$, $+0.232$] & $+0.145$ [$+0.095$, $+0.203$] \\
Bail $\beta=0.1$   & 0.126 & 0.252 & \textbf{0.396} & $+0.143$ [$+0.072$, $+0.207$] & $+0.115$ [$+0.015$, $+0.144$] \\
Bail $\beta=0.2$   & 0.251 & 0.517 & \textbf{0.661} & $+0.144$ [$+0.068$, $+0.258$] & $+0.153$ [$+0.024$, $+0.272$] \\
Credit $\beta=0.1$ & 0.228 & 0.352 & \textbf{0.644} & $+0.292$ [$+0.233$, $+0.376$] & $+0.249$ [$+0.179$, $+0.344$] \\
Credit $\beta=0.2$ & 0.260 & 0.539 & \textbf{0.782} & $+0.244$ [$+0.208$, $+0.302$] & $+0.201$ [$+0.132$, $+0.286$] \\
FICO $\beta=0.1$   & 0.118 & 0.576 & \textbf{0.841} & $+0.265$ [$+0.167$, $+0.302$] & $+0.211$ [$+0.129$, $+0.267$] \\
FICO $\beta=0.2$   & 0.200 & 0.747 & \textbf{0.941} & $+0.194$ [$+0.039$, $+0.215$] & $+0.121$ [$-0.024$, $+0.163$] \\
\midrule
German $\beta=0.1$ & 0.316 & 0.386 & \textbf{0.655} & $+0.269$ [$+0.069$, $+0.469$] & $+0.179$ [$-0.090$, $+0.446$] \\
German $\beta=0.2$ & 0.494 & 0.626 & \textbf{0.908} & $+0.282$ [$+0.090$, $+0.391$] & $+0.260$ [$-0.117$, $+0.348$] \\
HMEQ $\beta=0.1$   & 0.282 & 0.388 & \textbf{0.510} & $+0.121$ [$+0.090$, $+0.172$] & $+0.033$ [$-0.040$, $+0.130$] \\
HMEQ $\beta=0.2$   & 0.872 & 0.913 & \textbf{0.943} & $+0.030$ [$+0.000$, $+0.050$] & $+0.038$ [$+0.003$, $+0.065$] \\
\bottomrule
\end{tabular}
\end{table*}

\begin{figure*}[t]
\centering
\includegraphics[width=\linewidth]{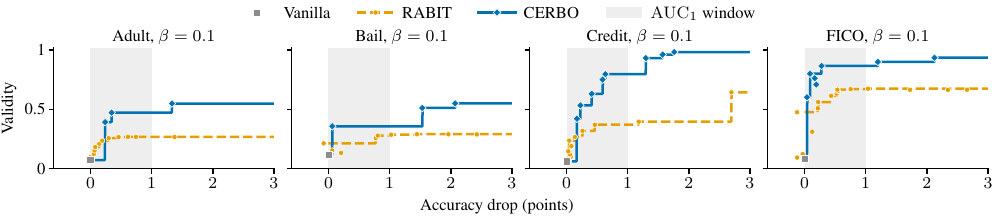}
\caption{Accuracy--recourse frontiers (validity) under additive costs ($\beta=0.1$, test
data, mean over seeds 200--207). Each marker is one trained model (RABIT: 18
weights; \ours: 8 requested rates), and lines are model frontiers, the best
validity of a trained model at each accuracy drop relative to Vanilla. The gray
band is the 1-point window of $\auc$ (Table~\ref{tab:main}). On all four data
sets \ours's frontier is at or above RABIT's at every accuracy drop of $0.25$
points or more. Appendix Figure~\ref{fig:frontiers-all} shows all 12 settings.}
\label{fig:frontiers}
\end{figure*}

\textbf{Protocol. }
We use RABIT's data, splits, cost and evaluation code without changes
\citep[Section~5 and App.~C]{kanamori2025rabit}: ten random 75/25 splits (seeds
300--309), the max percentile-shift cost with KDE percentiles, $\beta=0.2$, and
a search that samples 1{,}000 leaf regions. We add one
validation rule shared by both methods, described below. The metrics
are RABIT's: accuracy, the recourse ratio (the share of test individuals who are
accepted or for whom a valid action is found), the cost, sparsity and
plausibility of the found actions, ROC-AUC and F1.

Each recourse-aware method
gets one setting per data set, chosen by the same rule on validation data (an
80/20 split of the training data): the highest validation recourse ratio whose
validation accuracy is within 1 point of Vanilla's, the window of $\auc$. The
rule picks $\gamma=0.005$, $0.0035$, $0.004$ and $0.002$ from RABIT's published
grid $\{0.0005,\dots,0.005\}$ for FICO, COMPAS, Adult and Bail, and
$\rho=0.9$, $0.8$, $0.9$ and $0.5$ for \ours. We also report RABIT at its
published $\gamma=0.002$, which meets the rule on every data set. COMPAS has a
mutable one-hot feature outside our exactness results, so \ours trains on it
approximately.

\textbf{Results. }
Our runs of Vanilla and RABIT reproduce RABIT's published accuracy and recourse
ratio within the reported standard deviations on all four data sets, as
Table~\ref{tab:rabit} in the appendix shows. Table~\ref{tab:rabitbench} shows that \ours gives
recourse to more people, with cheaper actions and no significant loss of
accuracy. Paired over the ten splits against RABIT with tuned $\gamma$, the
accuracy difference is at most $0.001$ in absolute value on FICO, COMPAS and
Bail (none significant). The recourse ratio rises from $0.900$ to $0.992$, from
$0.966$ to $0.990$ and from $0.637$ to $0.769$ ($p\le10^{-4}$ each, paired
$t$-test), and the mean action cost falls by 18--54\%. The gains over
RABIT's published setting are at least as large. ROC-AUC is within $0.006$ of
both RABIT rows and F1 is equal or higher.

The comparison also holds over the whole trade-off. With all ten values of
RABIT's grid against all eight requested rates, \ours's $\auc$ on the recourse
ratio is larger by $+0.272$ [$+0.192$, $+0.327$] on FICO, $+0.067$ [$+0.009$,
$+0.092$] on COMPAS and $+0.146$ [$+0.073$, $+0.255$] on Bail. With
the selection-rule tolerances of Table~\ref{tab:tol}, from $0.5$ to $2$ points,
\ours's recourse ratio is never significantly lower. It leads significantly on
FICO and Bail, and on COMPAS the two methods tie at $0.5$ and $0.75$ points.

\ours's actions are the cheapest on every data set, their sparsity is within
$0.05$ of the best RABIT row, and their plausibility is worse than RABIT's only on
COMPAS ($0.491$ vs.\ $0.471$ and $0.486$; lower is better). Training times are of
the same order, as Table~\ref{tab:rabitsd} shows. On Adult \ours reaches a
recourse ratio of $0.999$, not counted because RABIT's KDE cost
prices small capital-gain increases as nearly free.

\textbf{Leaf refinement. }
RABIT's strongest variant refits leaf values after training. We apply its
released refinement to Vanilla, RABIT ($\gamma=0.002$) and \ours alike, at every level $\eta$ of
its grid ($0.01$ to $1.28$), and compare the methods at the same $\eta$, so no
level is selected. Figure~\ref{fig:rabitref} shows the result. Paired over the splits, \ours's
recourse ratio is higher at all eight levels on FICO, COMPAS and Bail. The
difference is significant ($p<0.05$) in 22 of 24 cases; the exceptions are
COMPAS at $\eta=0.32$ and $0.64$, where both methods exceed $0.98$. No accuracy
difference is significantly negative. Tables~\ref{tab:refgrid}
and~\ref{tab:t9rule} in the appendix list every level and apply RABIT's own
selection rule, under which \ours also leads.

% ---------------------------------------------------------------------------
\subsection{Frontiers under Additive Costs}
\label{sec:frontier}
Table~\ref{tab:main} and Figure~\ref{fig:frontiers} compare the frontiers under
additive costs. On the four
main data sets \ours raises $\auc$ over RABIT in all 8 settings, by $0.143$ to
$0.292$, with every 95\% interval above zero. On the held-out data sets the gain
is $0.030$ to $0.282$, and the intervals exclude zero except on HMEQ at
$\beta=0.2$, where both methods are near the ceiling and the lower end is
$0.000$.

On the stricter model frontier \ours is ahead in all 12 settings, with
the interval above zero in 7 of the 8 main settings (the exception is FICO at
$\beta=0.2$). The held-out intervals are wide (German has 1{,}000
individuals).

Figure~\ref{fig:frontiers} shows where the gain arises. At zero accuracy loss
RABIT can be ahead (FICO: $0.48$ vs.\ $0.08$), since every \ours model gives up a
little accuracy, but from a $0.25$-point drop on \ours leads on all four data
sets. At $0.5$ points its validity is $0.47$, $0.36$, $0.63$ and $0.87$ against
$0.27$, $0.21$, $0.37$ and $0.61$ on Adult, Bail, Credit and FICO. Beyond that,
RABIT's frontier gains at most $0.08$ on Adult, Bail and FICO, consistent with
the vacuous bound of Corollary~\ref{cor:nocert}, while \ours keeps rising. A
replication on four fresh seeds (700--703) gives the same picture: \ours
raises $\auc$ in all 12 settings, with the interval above zero except on
HMEQ, where the gap is below $0.01$.

RABIT's leaf refinement does not close the gap. Applied to every trained model
on Bail and FICO, under max and additive costs, it leaves \ours\,+\,refinement
ahead of the stronger refined baseline in all four settings, by $0.157$ to
$0.292$ in $\auc$, with every interval above zero (Table~\ref{tab:refine}). Nor
does a decision-threshold shift, RABIT's second post-processing: against all
Vanilla and RABIT models at eleven intercept shifts, \ours without a shift has the
larger $\auc$ in 11 of 12 settings, as Appendix~\ref{app:fr} details.

% ---------------------------------------------------------------------------
\subsection{Certified and Guaranteed Rates}
\label{sec:certexp}
\textbf{Certified recourse. }
We apply the test-time certifier of Theorem~\ref{thm:test} on fresh seeds and
all six data sets, measuring the share of rejected individuals for whom recourse
is proven. \ours raises the certified $\auc$ over RABIT in 11 of 12 settings, by
$0.08$ to $0.38$, with the interval above zero in 9 (Table~\ref{tab:cert}).
Requested rates also transfer to new people, as Figure~\ref{fig:weights}(b)
shows: for targets within 2 points of Vanilla's accuracy, the mean absolute gap
between $\rho$ and the test certified rate is $0.012$ to $0.090$.

\begin{table}[t]
\centering
\caption{Deploying with a guaranteed recourse rate (6 data sets $\times$ 2
budgets, 6 seeds $\times$ 200 calibration/evaluation splits, $\delta=0.1$,
Clopper--Pearson bound). Achievable: settings in which a qualifying model exists
in at least half of the deployments. Accuracy cost: Vanilla's accuracy minus the
deployed model's (median over settings achievable for both methods; lower is
better). Violation: share of deployments whose rate on new people falls short. Bold: the better method per witness; violations are not ranked. With either witness, \ours guarantees
higher rates at lower cost.}
\label{tab:control}
\small
\setlength{\tabcolsep}{4pt}
\begin{tabular}{lcccc}
\toprule
& \multicolumn{2}{c}{RABIT's search} & \multicolumn{2}{c}{Our certifier}\\
\cmidrule(lr){2-3}\cmidrule(lr){4-5}
Required rate & RABIT & \ours & RABIT & \ours \\
\midrule
30\% achievable & 11/12 & \textbf{12/12} & 7/12 & \textbf{12/12} \\
50\% achievable & 7/12 & \textbf{12/12} & 5/12 & \textbf{11/12} \\
70\% achievable & 2/12 & \textbf{11/12} & 2/12 & \textbf{9/12} \\
\midrule
Accuracy cost (points) & 0.66 & \textbf{0.17} & 1.35 & \textbf{0.33} \\
Violation & 0.8\% & 0.4\% & 1.2\% & 0.8\% \\
\bottomrule
\end{tabular}
\end{table}

\textbf{Guaranteed recourse rates. }
We simulate the recipe of Theorem~\ref{thm:control} on six data sets
($\beta\in\{0.1,0.2\}$, fresh seeds 800--805). Each method has a candidate
family: Vanilla plus \ours at eight requested rates, or Vanilla plus RABIT at
twelve values of $\gamma$. For each of 200 random splits of the test set into a
calibration and an evaluation half, we select the most accurate candidate whose
Clopper--Pearson bound meets the requirement ($\delta=0.1$) and evaluate it on
the other half. Table~\ref{tab:control} reports both witnesses, RABIT's search
and our certifier.

The guarantee holds: with RABIT's search as witness,
the rate on new individuals fell below the requirement in 0.4\% of 41{,}674
simulated deployments of \ours (9.0\% in the worst setting, within the allowed
10\%). Promises fail far less often than $\delta$ allows because the bound
covers all $M$ candidates at once, and this conservatism costs \ours little: a
median of 0.17 accuracy points relative to Vanilla, against 0.66 for RABIT.

\ours can guarantee a rate of 70\% in 11 of 12
settings and RABIT in 2. Where both can make a promise, \ours gives up less
accuracy in 18 of 20 settings. Guaranteeing 50\% on Adult at $\beta=0.2$ costs
0.2 points with \ours and 8.1 with RABIT. The selection rule is identical, so
the difference comes from the candidates: requested rates fill the gaps that,
by Proposition~\ref{prop:weights}, weights leave.

\begin{table}[t]
\centering
\caption{Ablation path under additive costs: $\auc$ on validity (envelope,
test data, seeds 200--207; higher is better). Each column adds one component to the
released RABIT (A): B, exact additive accounting in RABIT's training code; C,
exact split aggregation with RABIT's soft-min surrogate; D, convex midpoint
surrogate; E, exact additive reachability; F, committed boxes ($k=1$) at fixed
$\gamma$; G, targeting (\ours). The best value per row is in bold; targeting
gives the best frontier in 7 of 8 settings.}
\label{tab:abl}
\small
\setlength{\tabcolsep}{3.5pt}
\begin{tabular}{lccccccc}
\toprule
Setting & A & B & C & D & E & F & G \\
\midrule
Adult $0.1$  & 0.247 & 0.240 & 0.334 & 0.359 & 0.332 & 0.391 & \textbf{0.427} \\
Adult $0.2$  & 0.418 & 0.404 & 0.589 & \textbf{0.616} & 0.573 & 0.554 & 0.609 \\
Bail $0.1$   & 0.252 & 0.204 & 0.219 & 0.232 & 0.261 & 0.270 & \textbf{0.396} \\
Bail $0.2$   & 0.517 & 0.455 & 0.552 & 0.592 & 0.564 & 0.578 & \textbf{0.661} \\
Credit $0.1$ & 0.352 & 0.324 & 0.305 & 0.354 & 0.629 & 0.576 & \textbf{0.644} \\
Credit $0.2$ & 0.539 & 0.525 & 0.701 & 0.765 & 0.746 & 0.654 & \textbf{0.782} \\
FICO $0.1$   & 0.576 & 0.576 & 0.601 & 0.701 & 0.708 & 0.729 & \textbf{0.841} \\
FICO $0.2$   & 0.747 & 0.675 & 0.840 & 0.891 & 0.908 & 0.761 & \textbf{0.941} \\
\bottomrule
\end{tabular}
\end{table}

% ---------------------------------------------------------------------------
\subsection{Efficiency and Deployed Ensembles}
\label{sec:libs}
\textbf{Training and certification time. }
On Adult with 2{,}000 to 32{,}000 training rows (one run at a time), \ours trains
in $1.3\times$ RABIT's time at full size (201\,s vs.\ 156\,s; at most $1.9\times$,
at the smallest size). Its training time grows near-linearly (log--log slope
$1.16$; RABIT $1.31$), as Corollary~\ref{cor:agg} suggests, and certification
takes under 1\,ms per person at every size. In the certification runs, the
certifier takes a median of 0.6--1.9\,ms per person and RABIT's search
2.2--7.0\,ms, except on German with 51 features (about 10\,ms each); the exact
program takes 0.9--4.1\,s.

\textbf{Certifying XGBoost and LightGBM. }
Theorem~\ref{thm:test} applies to any tree ensemble. We convert XGBoost and
LightGBM models (up to 1{,}000 trees with early stopping; seeds 400--404) into the certifier's format. On every test point their decisions match the
libraries', and all certificates in the 120 runs are sound when checked against
the original library model. Certification takes a median of 7.8\,ms per person, against
41.6\,ms for RABIT's search. These models, not trained for recourse, offer
little of it: search finds verified actions for 9--73\% of rejected people, and
the certifier certifies 0.1--23\%. At similar accuracy \ours offers several times
more; on FICO at $\beta=0.1$, (accuracy, validity) is $(0.728,0.09)$ for XGBoost
and $(0.719,0.91)$ for \ours. The rate-control recipe can still audit such a
model: with RABIT's search as witness, the bound failed in 8.5\% of calibration
draws ($\delta=0.1$).

% ---------------------------------------------------------------------------
\subsection{Ablations and Analysis}
\label{sec:ablate}

\textbf{Ablations. }
Table~\ref{tab:abl} adds one component at a time to RABIT. Exact additive
accounting inside RABIT's training code (B) does not help by itself. Exact split
aggregation with RABIT's soft-min (C) improves on RABIT in 6 of 8 settings, and
the convex surrogate (D) improves on C in every setting, so the gain is not due
to the aggregation alone. Exact reachability (E) and committed boxes (F) help in some settings and not in
others, and targeting (G, \ours) gives the best frontier in 7 of 8. Against D, \ours
is ahead in 7 of 8 settings, with the interval above zero in 5. No fixed-$\gamma$
variant matches the full method consistently, which is the practical case for
targeting: one run per requested rate instead of a dense $\gamma$ search.

\textbf{Exact recourse. }
Because search can miss actions, we decide recourse exactly for 100 rejected
test people per model on
Adult, Bail, Credit and FICO ($\beta\in\{0.1,0.2\}$, seeds 200--202; 384 models,
35{,}009 people), using RABIT's mixed-integer formulation plus the budget
constraint, solved by SCIP. One person hit the 20\,s limit and counts as having no recourse. On exact validity \ours
leads RABIT in all 8 settings and all 24 setting--seed pairs, by $0.115$--$0.364$
$\auc$. Exact and search-based gaps agree within $0.025$. On \ours's models,
0.79 of the decided people have recourse, search finds 0.77 and the certifier
0.57.

\textbf{When committed boxes help. }
On synthetic data in which the label depends equally on $m$ mutable features
(16 regimes $\times$ 2 budgets, seeds 10--13), committed boxes help over exact
reachability alone when few features are mutable ($m=2$: $\auc$ $0.348$ vs.\
$0.306$) and hurt when many are ($m\ge6$: $-0.08$ to $-0.10$), because a box
commits to directions early, as Proposition~\ref{prop:kmono} suggests. \ours
inherits this behavior, and Appendix~\ref{app:rob} shows that choosing the
commitment depth on validation data does not remove it. The real data sets have
4--15 mutable features, and on all of them \ours leads RABIT.

% ---------------------------------------------------------------------------
\section{Related Work}
\textbf{Algorithmic recourse. }
\citet{karimi2022survey} survey recourse and counterfactual explanations.
Post-hoc methods search for actions in a fixed model,
for linear models \citep{ustun2019} or model-agnostically
\citep{wachter2018,karimi2020,mothilal2020}. For tree ensembles they use feature
tweaking \citep{tolomei2017} or mixed-integer optimization
\citep{parmentier2021}, and counterfactual explanation trees assign consistent
actions to groups \citep{kanamori2022}. CARLA benchmarks such
methods \citep{pawelczyk2021carla}, reachable-set verification decides whether
any action exists \citep{kothari2024}, and robust recourse methods guard against
model and input changes \citep{upadhyay2021robust,dominguez2022robust}. These
methods explain a model after training; \ours changes the model so that recourse
exists.

\textbf{Recourse-aware learning. }
Other methods shape the model itself: adversarial training with PAC
confidence sets \citep{ross2021}, robust recourse models
\citep{guo2023}, recourse-aware trees and forests \citep{kanamori2024}, and
RABIT for GBDTs \citep{kanamori2025rabit}. Relative to RABIT, \ours adds exact additive training, certificates and rate
guarantees.

\textbf{Verification of tree ensembles. }
Committed boxes relate to robustness verification of tree ensembles over the
regions reachable under a perturbation budget
\citep{kantchelian2016,chen2019robustness,einziger2019verigb} and to provably
robust boosting against a worst-case bound \citep{andriushchenko2019}. Recourse reverses the quantifier: it needs
\emph{some} affordable change to be approved, not every change to be rejected,
so by Theorem~\ref{thm:sandwich} a worst-case bound like RABIT's is vacuous for
rejected individuals; committed boxes turn it into a certificate.

\textbf{Statistical guarantees for deployed models. }
In machine learning systems, ease.ml/ci gives $(\epsilon,\delta)$ guarantees on
model-quality conditions from test sets in continuous integration
\citep{renggli2019easeml}; Section~\ref{sec:deploy} certifies a recourse rate in
the same way.

% ---------------------------------------------------------------------------
\section{Limitations}
Our guarantees need coordinate-separable feasible sets and per-feature costs
monotone in the distance moved. Grouped categorical or causal
actions would need grouped reachability states. Certificates are sound but
incomplete, and validity relies on an incomplete search, a lower bound on true
recourse that Section~\ref{sec:ablate} shows to be close.
Actionability metadata determines which actions count; with the stricter Adult
metadata of Appendix~\ref{app:rob}, every method's validity drops but \ours still
leads. As Section~\ref{sec:ablate} shows, committed boxes lose recourse when the
label depends on small joint moves across many mutable features; a per-person
commitment rule remains open.

The rate guarantee assumes calibration and deployment individuals come from one
distribution, and it
covers the population rate: it promises recourse neither to every rejected
person nor equally across groups, and per-group rates would need calibration
data from each group. Because the bound covers all $M$ candidates at once, it is
conservative; promises failed in 0.4\% of deployments where $\delta$ allows 10\%,
and a smaller calibration sample makes high rates harder to guarantee. Finally,
an action that reverses the decision need not improve the outcome the model
predicts. Appendix~\ref{app:impact} discusses broader impact.

% ---------------------------------------------------------------------------
\section{Conclusion}
A single recourse weight cannot deliver a requested rate, and RABIT's training
bound cannot certify recourse for any rejected individual. \ours replaces the weight with a request,
certifies actions and guarantees the rate on new people. In both settings it
gives recourse to more people than RABIT at equal accuracy and with cheaper
actions, and its certifier can audit any deployed tree ensemble. An operator
can thus state a recourse rate in advance, verify it on held-out applicants and
re-check it with the certifier whenever the applicant population changes.

\bibliographystyle{mlsys2025}
\bibliography{refs}

% ---------------------------------------------------------------------------
% Appendix
\clearpage
\onecolumn
% Appendix: one column, so that floats sit next to their text.
\makeatletter
\setlength{\@fptop}{0pt}\setlength{\@fpbot}{0pt plus 1fil}
\setlength{\@fpsep}{12pt plus 2pt minus 4pt}
\makeatother
\setlength{\floatsep}{12pt plus 2pt minus 4pt}
\appendix
\section{Proofs}\label{app:proofs}

\begin{lemma}[Intersection cost]\label{lem:max}
If $I$ is an interval, $H$ a half-line and $I\cap H\neq\emptyset$, then
$\kappa(x;I\cap H)=\max(\kappa(x;I),\kappa(x;H))$ for monotone $\phi$.
\end{lemma}
\begin{proof}
For monotone $\phi$ the cheapest point of an interval lies at $x$ itself or at
the endpoint on the side of $x$. The binding endpoint of $I\cap H$ on that side
is the endpoint of $I$ or of $H$ that is farther from $x$, whose cost is the
larger of the two.
\end{proof}

\begin{proof}[Proof of Proposition~\ref{prop:box}]
Any feasible $a$ with $x+a\in B$ satisfies $x_d+a_d\in I_d$ for all $d$, so
$c(a\mid x)\ge\sum_d\kappa_d(x;I_d)$. Coordinate-wise minimizers combine into a
feasible action because $\mathcal A(x)$ and $c$ are separable.
\end{proof}

\begin{proof}[Proof of Proposition~\ref{prop:caps}]
A product set containing the vertices $\beta e_1$ and $\beta e_2$ must have
$b_1,b_2\ge\beta$ and so contains $\beta(e_1+e_2)$, whose sum is $2\beta>\beta$.
\end{proof}

\begin{proof}[Proof of Proposition~\ref{prop:mono}]
For $\tau_q<x_{id}$ the cost of crossing to the left decreases as $\tau_q$ approaches
$x_{id}$; for $\tau_q\ge x_{id}$ the cost of crossing to the right increases with
$\tau_q$. So reaching the left child is non-decreasing in $q$ and reaching the
right child is non-increasing. If the left child is not reachable then
$\tau_q<x_{id}$ and the sample lies on the right.
\end{proof}

\begin{proof}[Proof of Corollary~\ref{cor:agg}]
By Proposition~\ref{prop:mono} each sample contributes to one contiguous range
of thresholds for each of the three sets. Its two transition indices are found
by binary search ($O(\log K)$ each); adding its gradient and Hessian at the
range ends of three difference arrays and taking prefix sums gives all $K$
sums in $O(r\log K+K)$.
\end{proof}

\begin{proof}[Proof of Lemma~\ref{lem:mid}]
Since $\max(w_L,w_R)\ge(w_L+w_R)/2$ and $\ell$ is non-increasing,
$\min_j\ell(s+w_j)=\ell(s+\max_jw_j)\le\ell(s+(w_L+w_R)/2)$. A convex function
of a linear map is convex, and the Hessian follows from the chain rule.
\end{proof}

\begin{lemma}[Indefinite soft-min model]\label{lem:indef}
Let $\ell(s)=\log(1+e^{-s})$, $\nu>0$, and
$M=\frac1\nu\begin{psmallmatrix}\ell''(s)/2&\ell'(s)^2/4\\\ell'(s)^2/4&\ell''(s)/2\end{psmallmatrix}$,
the two-leaf second-order model of RABIT's soft-min at $w=0$ as implemented in
its released code. Then $M$ is indefinite if and only if $\sigma(s)<1/3$.
\end{lemma}
\begin{proof}
The eigenvalues of $M$ are $\frac1\nu\big(\ell''/2\pm\ell'^2/4\big)$. With
$\ell'=-(1-\sigma)$ and $\ell''=\sigma(1-\sigma)$, the larger one is positive, and
the smaller one equals $(1-\sigma)(3\sigma-1)/(4\nu)$, which is negative exactly
when $\sigma<1/3$.
\end{proof}

\begin{proof}[Proof of Theorem~\ref{thm:valid}]
Initially $F_0=0=s_i$ and $B_i=\R^D$. Suppose \eqref{eq:inv} holds before tree
$t$, and let $a\in\mathcal A_\beta(x_i)$ with $x_i+a\in B_i\cap R_u$. Then $x_i+a$
lies in a leaf $v$ below $u$ that is reachable within the old box and budget,
so $f_t(x_i+a)\ge m_i(u)$, and $F_{t-1}(x_i+a)\ge s_i$ by hypothesis. Adding
gives $F_t(x_i+a)\ge s_i+m_i(u)$. The loss bound follows because $\ell$ is
non-increasing and the minimum over a subset of $\mathcal A_\beta(x_i)$ is at
least the minimum over the whole set. For $k=0$ the node is the root, so the box
is unchanged and $m_i$ is the minimum reachable leaf value, which is RABIT's
per-tree term. With leaf commitment every committed box lies inside one leaf of
each tree, so $F_{t-1}$ is constant on $B_i$ and equals $s_i$.
\end{proof}

\begin{proof}[Proof of Theorem~\ref{thm:cert}]
The projection $a_i$ attains $C(x_i,B_i)\le\beta$ by Proposition~\ref{prop:box}
(commitment only accepts reachable nodes, and emptiness checks respect integer
features), so $a_i\in\mathcal A_\beta(x_i)$ and $x_i+a_i\in B_i$; then
\eqref{eq:inv} gives $F_T(x_i+a_i)\ge s_i$.
\end{proof}

\begin{proof}[Proof of Theorem~\ref{thm:test}]
The proof of Theorem~\ref{thm:valid} uses only the tree structure and leaf
values, not how the trees were fitted, and the proof of Theorem~\ref{thm:cert}
then applies. Each tree costs $O(2^kD)$: the reachable nodes at depth $k$ are
examined with $O(D)$ cost updates each.
\end{proof}

\begin{proof}[Proof of Theorem~\ref{thm:sandwich}]
\emph{Left chain.} For every $a\in\mathcal A_\beta(x)$ and every $t$,
$\min_{a'\in\mathcal A_\beta(x)}f_t(x+a')\le f_t(x+a)$; summing over $t$ gives
$\xi(x)\le F(x+a)$, hence $\xi(x)\le\min_aF(x+a)$. Since $a=0$ is feasible with
cost $0\le\beta$, $\min_aF(x+a)\le F(x)$.

\emph{$s_0=\xi$.} With $k=0$ every commitment is to the root, the box stays
$\R^D$, and the credit added for tree $t$ is the minimum value over the leaves of
$f_t$ reachable from $x$ within budget, which is
$\min_{a\in\mathcal A_\beta(x)}f_t(x+a)$.

\emph{$s_k\ge\xi$.} Consider tree $t$ and the current box $B$ (affordable: some
$a\in\mathcal A_\beta(x)$ has $x+a\in B$). The leaves reachable from $x$ within
$B$ and budget form a subset of the leaves reachable within budget alone, and
this subset is non-empty because the leaves partition $\R^D$. Let $u$ be the
committed node and $u_0$ the depth-$k$ ancestor of any reachable leaf. Then the
credit $m(u)\ge m(u_0)\ge\min\{\text{values of leaves reachable within }B\}\ge
\min\{\text{values of leaves reachable within budget}\}$, the last quantity
being RABIT's term for tree $t$. Summing over $t$ gives $s_k(x)\ge\xi(x)$.

\emph{$s_k\le\max$.} By Theorem~\ref{thm:test} the projection $a^\star$ of $x$
onto the final box is in $\mathcal A_\beta(x)$ and satisfies
$F(x+a^\star)\ge s_k(x)$.
\end{proof}

\begin{proof}[Proof of Corollary~\ref{cor:nocert}]
Immediate from $\xi(x)\le F(x)$ in Theorem~\ref{thm:sandwich}. Clipping each
term at $0$, as in RABIT's implementation, only decreases $\xi$.
\end{proof}

\begin{proof}[Proof of Proposition~\ref{prop:gap}]
Take one mutable feature, let every tree be the stump
$f_t(z)=w$ for $z>0$ and $f_t(z)=-w$ for $z\le0$, and let $x<0$ be close enough to
$0$ that crossing to $z>0$ costs at most $\beta$. Both leaves of every tree are
reachable, so $\xi(x)=-Tw=F(x)$. For $k=1$ the leaves are the depth-1 nodes: at
tree 1 the sample commits to the right leaf (credit $w$, box $z>0$), after which
only the right leaf of every later tree is reachable within the box, so
$s_1(x)=Tw=\max_{a}F(x+a)$.
\end{proof}

\begin{proof}[Proof of Proposition~\ref{prop:weights}]
(i) If $F_\gamma$ minimizes $L+\gamma\,\Omega$, then
$L(F)+\gamma\,\Omega(F)\ge L(F_\gamma)+\gamma\,\Omega(F_\gamma)$ for every $F\in\mathcal F$, so
the line through $(L(F_\gamma),\Omega(F_\gamma))$ with slope $-1/\gamma$ (vertical for
$\gamma=0$) has all of $\mathcal P$ on or above it. A supported point of $\mathcal P$ is on the lower
convex hull; a point strictly above the hull admits no supporting line and is
never a minimizer.

(ii) Let $\gamma_1<\gamma_2$ with minimizers $F_1,F_2$. Optimality gives
$L_1+\gamma_1\Omega_1\le L_2+\gamma_1\Omega_2$ and $L_2+\gamma_2\Omega_2\le L_1+\gamma_2\Omega_1$. Adding
the two inequalities gives $(\gamma_2-\gamma_1)(\Omega_2-\Omega_1)\le0$, so $\Omega_2\le\Omega_1$, and
then the first inequality gives $L_1\le L_2$. By (i) the minimizer is a hull
vertex, and it can only change at the slopes of the hull's edges.

(iii) Let $\mathcal F=\{A,B,C\}$ with $(L,\Omega,V)$ equal to $(0.10,1,0)$,
$(0.16,0.5,0.5)$ and $(0.20,0,1)$, where $\Omega=1-V$. Then $L+\gamma\,\Omega$ equals
$0.10+\gamma$, $0.16+0.5\gamma$ and $0.20$ respectively. $B$ beats $A$ only for
$\gamma>0.12$ and beats $C$ only for $\gamma<0.08$, so $B$ is never returned. For
$\gamma<0.1$ the minimizer is $A$ (rate $0$), and for $\gamma>0.1$ it is $C$
(rate $1$, loss $0.20>0.16$). For $\rho\in(0,0.5]$ the solution of
\eqref{eq:req} is $B$.
\end{proof}

\begin{proof}[Proof of Proposition~\ref{prop:track}]
Taking logarithms, $\log\gamma_{T+1}-\log\gamma_1=\lambda\sum_t(\rho-\hat V_t)$, and both
logarithms lie in $[\log\gamma_{\min},\log\gamma_{\max}]$.
\end{proof}

\begin{proof}[Proof of Theorem~\ref{thm:control}]
The models are fixed with respect to $S$. Fix $j$ and condition on the set of
calibration individuals that $F_j$ rejects; given $n_j$, these are i.i.d.\
draws from $P(\cdot\mid F_j(x)\le0)$, and the witness is a fixed
$\{0,1\}$-valued function of $x$ with mean $V(F_j)$, so $z_j$ is binomial with
parameters $n_j$ and $V(F_j)$. Hoeffding's inequality, or the defining property
of the Clopper--Pearson bound, gives $\Pr[V(F_j)<\underline V_j]\le\delta/M$, and a union
bound gives $\Pr[\exists j: V(F_j)<\underline V_j]\le\delta$. On the complementary event,
$V(F_J)\ge\underline V_J\ge\rho$ for whichever $J$ the rule selects, because the event
holds for all $j$ simultaneously. The second claim holds because the witness is
sound (for the certifier, Theorem~\ref{thm:test}).
\end{proof}

\begin{proposition}[Credit is monotone in $k$]\label{prop:kmono}
For a fixed tree and box, the best achievable credit $\max_um_i(u)$ over nodes
of depth $k$ is non-decreasing in $k$.
\end{proposition}
\begin{proof}
For a node $u$ with children $v,w$,
$m_i(u)=\min(m_i(v),m_i(w))\le\max(m_i(v),m_i(w))$.
\end{proof}

\begin{proposition}[Starved window]\label{prop:starved}
Under RABIT's depth-scheduled cap $\beta((q+1)/D_{\max})^{0.3}$ at depth $q$ of a
tree of depth $D_{\max}$, a child whose only within-budget access is one
crossing of cost in $(\beta((q+1)/D_{\max})^{0.3},\beta]$ is marked unreachable
although it is reachable. At the root with $D_{\max}=8$ the window is $(0.536\beta,\beta]$.
\end{proposition}
\begin{proof}
Immediate from the definitions.
\end{proof}

% ---------------------------------------------------------------------------
\section{Implementation Notes}\label{app:impl}
Algorithm~\ref{alg:cerbo} lists the training loop of \ours and
Algorithm~\ref{alg:certify} the test-time certifier; both use the commitment step
of Section~\ref{sec:cb}, implemented as described in Section~\ref{sec:impl}.

\begin{algorithm}[!htb]
\caption{\ours training at a requested rate $\rho$}
\label{alg:cerbo}
\begin{algorithmic}[1]
\REQUIRE training data $\{(x_i,y_i)\}_{i=1}^n$, feasible sets $\mathcal A$, cost
$c$, budget $\beta$, rate $\rho$, trees $T$, depth $k$, $\gamma_1$, $\lambda$,
$[\gamma_{\min},\gamma_{\max}]$
\STATE $F_0\leftarrow0$; \ $B_i\leftarrow\R^D$, $s_i\leftarrow0$ for all $i$
\FOR{$t=1$ {\bfseries to} $T$}
\STATE grow $f_t$ greedily on the loss plus $\gamma_t$ times the recourse bound of
rejected samples, using the credit $s_i$, exact reachability of $B_i$
within budget $\beta$ (Section~\ref{sec:exact}) and the midpoint bound
(Lemma~\ref{lem:mid})
\STATE $F_t\leftarrow F_{t-1}+f_t$
\FOR{$i=1$ {\bfseries to} $n$}
\STATE choose the node $u$ of $f_t$ at depth $k$ (or a shallower leaf) with
$R_u$ reachable from $B_i$ within budget that maximizes $m_i(u)$
\STATE $B_i\leftarrow B_i\cap R_u$; \ $s_i\leftarrow s_i+m_i(u)$
\ENDFOR
\STATE $\hat V_t\leftarrow$ share of rejected samples ($F_t(x_i)\le0$) with $s_i>0$
\STATE $\gamma_{t+1}\leftarrow\gamma_t\exp(\lambda(\rho-\hat V_t))$, projected onto $[\gamma_{\min},\gamma_{\max}]$
\ENDFOR
\STATE {\bfseries return} $F_T$ and, for each $i$ with $s_i>0$, the certified
action $a_i$ that projects $x_i$ onto $B_i$
\end{algorithmic}
\end{algorithm}

\begin{algorithm}[!htb]
\caption{Test-time certifier for any tree ensemble (Theorem~\ref{thm:test})}
\label{alg:certify}
\begin{algorithmic}[1]
\REQUIRE trees $f_1,\dots,f_T$, input $x$, feasible set $\mathcal A(x)$, cost
$c$, budget $\beta$, depth $k$
\STATE $B\leftarrow\R^D$; \ $s\leftarrow0$
\FOR{$t=1$ {\bfseries to} $T$}
\STATE choose the node $u$ of $f_t$ at depth $k$ (or a shallower leaf) with
$R_u$ reachable from $B$ within budget that maximizes $m(u)$
\STATE $B\leftarrow B\cap R_u$; \ $s\leftarrow s+m(u)$
\ENDFOR
\STATE {\bfseries if} $s>0$ {\bfseries return} the action $a$ that projects
$x$ onto $B$; it satisfies $F(x+a)\ge s>0$ and $c(a\mid x)\le\beta$
\STATE {\bfseries else return} ``not certified'' (recourse may still exist)
\end{algorithmic}
\end{algorithm}

\textbf{Hardware. }
Experiments ran on Kaggle CPU sessions (4 vCPUs of an Intel Xeon at 2.20\,GHz,
30\,GB RAM, Ubuntu; four models in parallel) and, for part of the frontier and
certification runs, on a local machine (Intel Core i7-12650H, 10 cores, 16\,GB RAM,
Windows 11; ten models in parallel). Training-time comparisons use Kaggle only.

\textbf{Software and licenses. }
We pin Python 3.11, NumPy 1.26.4 (BSD-3-Clause), SciPy 1.14.1
(BSD-3-Clause), pandas 2.2.3 (BSD-3-Clause), scikit-learn 1.5.2 (BSD-3-Clause),
Numba 0.60.0 (BSD-2-Clause), XGBoost 2.1.3 (Apache-2.0), LightGBM 4.5.0 (MIT) and
PySCIPOpt 5.5.0 (MIT) with SCIP (Apache-2.0). All comparisons use RABIT's public
code (\url{https://github.com/kelicht/rabit}) at commit \texttt{0cf0068} under its
BSD-3-Clause license, and our runs of its released configuration are
bit-identical to it.

\textbf{Differences from RABIT's code. }
Our exact aggregation differs from the
public implementation in four places that matter under per-sample budgets:
(i) the both-reachable set is built from the transitions of
Proposition~\ref{prop:mono} rather than from a sample-sorted scan; (ii) the
both-reachable gradient enters with the sign of the bound it expands; (iii) the
recourse bound $\xi$ is indexed by the original row of each bootstrap-sampled
training point; and (iv) integer crossings are rounded to the feature's integer
grid. All methods share the action search, data splits and metric code, so
these choices affect only the training objective; ablation B
isolates exact accounting alone. Automated tests verify the baseline
reproduction, the exact aggregation against a direct evaluation of the
reachability sets, and the soundness of every certificate.

\section{Additional Results}\label{app:extra}
Appendix~\ref{app:setup} lists the data sets, the settings and the runs behind
every result. The other subsections follow Section~\ref{sec:exp}: Appendix~\ref{app:rb}
covers RABIT's protocol, Appendix~\ref{app:fr} the frontiers under additive costs,
Appendix~\ref{app:cr} certified and guaranteed rates, and Appendix~\ref{app:rob}
robustness checks.

\subsection{Data Sets, Settings and Runs}\label{app:setup}
Table~\ref{tab:data} describes the seven data sets, Table~\ref{tab:settings}
lists the settings of every method, and Table~\ref{tab:runs} maps each result to
its data sets, seeds and runs.

\begin{table}[!htb]
\centering
\caption{Data sets. $N$: individuals; $D$: features after one-hot encoding;
mutable: encoded features that an action may change (RABIT's actionability
metadata for the first five data sets, ours for German and HMEQ); favorable: share
of individuals with the favorable label. German and HMEQ, below the rule, were
added after method development. All seven are public, and licenses are as stated
by each provider (HMEQ: Kaggle data set \texttt{ajay1735/hmeq-data}).
COMPAS is RABIT's released file (its paper lists $6{,}167\times14$).}
\label{tab:data}
\small
\hypersetup{linkcolor=black,citecolor=black}
\begin{tabular}{lrrrcll}
\toprule
Data set & $N$ & $D$ & Mutable & Favorable & Source & License\\
\midrule
FICO   & 9{,}871  & 23 & 15 & 0.480 & \citet{fico2018} & FICO terms\\
COMPAS & 7{,}214  & 15 & 7  & 0.549 & \citet{angwin2016} & none stated\\
Adult  & 48{,}842 & 16 & 5  & 0.239 & \citet{becker1996adult} & CC BY 4.0\\
Bail   & 8{,}923  & 16 & 7  & 0.375 & \citet{schmidt1988} & ICPSR terms\\
Credit & 30{,}000 & 13 & 9  & 0.779 & \citet{yeh2009} & CC BY 4.0\\
\midrule
German & 1{,}000  & 51 & 7  & 0.700 & \citet{hofmann1994} & CC BY 4.0\\
HMEQ   & 5{,}960  & 18 & 6  & 0.801 & \citet{baesens2016} & CC0 1.0\\
\bottomrule
\end{tabular}
\end{table}

\begin{table}[!htb]
\centering
\caption{Settings of every method. Each method spans its grid of weights or
requested rates; Section~\ref{sec:rabitbench}
selects one value per method on validation data, and the frontiers use all of
them. Proposition~\ref{prop:starved} analyzes RABIT's depth-scheduled budget
rule.}
\label{tab:settings}
\small
\begin{tabular}{@{}l>{\raggedright\arraybackslash}p{0.76\textwidth}@{}}
\toprule
Component & Setting\\
\midrule
Shared booster & RABIT's implementation and configuration for every method: 100 trees of depth 8, learning rate $0.1$, $\sqrt D$ candidate features per split, 128 bins, ridge term $\alpha=0.001$\\
Splits & per seed, a stratified 75/25 train/test split; validation data, where used, are a stratified 80/20 split of the training data\\
RABIT & soft-min surrogate with $\nu=1$; under additive costs, the depth-scheduled budget rule of its code. Weights: its published grid $\{0.0005,0.001,\dots,0.005\}$ under RABIT's protocol; 12 weights $\{0.001, 0.002, 0.003, 0.005, 0.007, 0.01, 0.015, 0.02, 0.03, 0.05, 0.1, 0.2\}$ elsewhere, plus $\{0.0012, 0.0014, 0.0016, 0.0018, 0.0025, 0.004\}$ for the 18-weight frontiers\\
\ours & requested rates $\rho\in\{0.2,0.3,\dots,0.9\}$; commitment depth $k=1$; midpoint surrogate; initial $\gamma=0.01$, $\lambda=0.5$, $[\gamma_{\min},\gamma_{\max}]=[10^{-3},1]$\\
Ablations & B and C: RABIT's 12 weights; D, E and F: 17 weights from $0.005$ to $1$\\
Leaf refinement & RABIT's released code at $\eta\in\{0.01,0.02,0.04,\dots,1.28\}$, calibration samples from the training split\\
Validity & RABIT's action search over 1{,}000 sampled leaf regions, in chunks of 512 individuals, for every method\\
Exact decisions & RABIT's mixed-integer formulation plus the budget constraint, SCIP, 20\,s limit per person, 100 rejected test people per model\\
Rate control & Clopper--Pearson bound at level $\delta/M$ over $M$ candidates, $\delta=0.1$, 200 calibration/evaluation halves per seed\\
XGBoost & up to 1{,}000 trees, learning rate $0.05$, depth 6, row and column subsampling $0.8$, histogram method; early stopping after 30 rounds without improvement on a stratified 15\% split of the training data\\
LightGBM & as XGBoost, with 31 leaves per tree instead of a depth limit\\
Accuracy reference & scikit-learn's histogram gradient boosting, default and tuned (up to 1{,}000 iterations, learning rate $0.05$, early stopping as above)\\
\bottomrule
\end{tabular}
\end{table}

\begin{table}[!htb]
\centering
\caption{The runs behind every reported number. Each seed is a fresh split,
unused in method development (Section~\ref{sec:setup}). Runs: number of
trained models, including those trained on validation splits for
model selection. CPU-h: summed single-core task time. Method development (seeds
0--12 and 100--107) used further compute, not counted here. Budgets are for the
additive cost unless marked max. RABIT's four: FICO, COMPAS, Adult and Bail;
four main: Adult, Bail, Credit and FICO; six: these and German and HMEQ. The
synthetic data have 16 regimes; training time uses subsets of Adult with
2{,}000 to 32{,}000 rows.}
\label{tab:runs}
\small
\hypersetup{linkcolor=black,citecolor=black}
\setlength{\tabcolsep}{4pt}
\begin{tabular}{@{}llllrrr@{}}
\toprule
Experiment & Sections & Data sets & $\beta$ & Seeds & Runs & CPU-h\\
\midrule
RABIT's protocol & \ref{sec:rabitbench}, \ref{app:rb} & RABIT's four & 0.2 (max) & 300--309 & 1{,}600 & 89\\
Frontiers and ablations & \ref{sec:frontier}, \ref{sec:ablate}, \ref{app:fr} & six & 0.1, 0.2 & 200--207 & 9{,}489 & 341\\
Refinement along the trade-off & \ref{sec:frontier}, \ref{app:fr} & Bail, FICO & 0.2 (max); 0.1 & 300--309; 200--204 & 330 & 37\\
Replication, commitment depth & \ref{sec:frontier}, \ref{app:rob} & six; synthetic & 0.1, 0.2 & 700--703; 20--22 & 6{,}623 & 220\\
Certified recourse & \ref{sec:certexp}, \ref{app:cr} & six & 0.1, 0.2 & 400--404 & 2{,}278 & 76\\
Guaranteed rates & \ref{sec:certexp} & six & 0.1, 0.2 & 800--805 & 1{,}511 & 47\\
Threshold shifts, reference & \ref{sec:frontier}, \ref{app:fr}, \ref{app:rob} & six & 0.1, 0.2 & 900--904 & 1{,}377 & 43\\
XGBoost and LightGBM & \ref{sec:libs} & six & 0.1, 0.2 & 400--404 & 120 & 8\\
Training time & \ref{sec:libs} & Adult (subsets) & 0.1 & 600--602 & 45 & 1\\
Exact recourse & \ref{sec:ablate} & four main & 0.1, 0.2 & 200--202 & 384 & 36\\
Synthetic data & \ref{sec:ablate} & synthetic & 0.1, 0.2 & 10--13 & 3{,}456 & 136\\
Base-model accuracy & \ref{sec:setup}, \ref{app:rob} & Bail & 0.1, 0.2 & 1000--1004 & 695 & 13\\
Stricter actionability & \ref{app:rob} & Adult (stricter) & 0.1, 0.2 & 500--504 & 220 & 17\\
\midrule
Total & & & & & 28{,}128 & 1{,}064\\
\bottomrule
\end{tabular}
\end{table}

\FloatBarrier
\subsection{Comparison under RABIT's Protocol}\label{app:rb}
\textbf{Correspondence to RABIT's experiments. }
Section~\ref{sec:rabitbench} adds \ours to the experiments of
\citet{kanamori2025rabit} as follows. Their baseline comparison (Fig.~1,
Table~1 and Fig.~5, with standard deviations in Table~6) corresponds to
Table~\ref{tab:rabitbench}, with every metric and its standard deviation in
Table~\ref{tab:rabitsd}; their running times (Fig.~2) are in
Table~\ref{tab:rabitsd}; and their leaf refinement (Fig.~4 and Table~9)
corresponds to Figure~\ref{fig:rabitref} and Tables~\ref{tab:refgrid}
and~\ref{tab:t9rule}. Three of their analyses have counterparts in another
setting: their sensitivity to $\gamma$ (Fig.~3) corresponds to RABIT's full
$\gamma$ grid in Section~\ref{sec:rabitbench} and the 18-weight frontiers of
Section~\ref{sec:frontier}; their exact action search (Table~8) to our exact
decisions (Section~\ref{sec:ablate}); and their intercept adjustment (Fig.~8) to
our decision-threshold shifts (Section~\ref{sec:frontier}). We do not repeat
their sensitivity to the number of trees and the budget (Figs.~6 and 7) or their
brittleness analysis (Table~7).

\textbf{Protocol differences. }
We follow their protocol with four changes:
(i) our ten splits use seeds 300--309, so every method, RABIT included, is rerun
on the same splits; (ii) the action search runs in chunks of 512 individuals and
draws its 1{,}000 leaf regions per chunk, for every method alike; (iii) their
Appendix~B.2 describes enumerating all leaves, whereas the released code, which
produced their numbers and which we use, samples 1{,}000 regions; and (iv)
refinement draws its calibration samples from the training data, as in the
released code for their Table~9, whereas their Fig.~4 used a separate 50:25:25
calibration split, which we did not rerun.

\textbf{Reproduction. }
Table~\ref{tab:rabit} compares our runs of RABIT's released configuration with
its published values: accuracy and the recourse ratio agree within $0.01$, and
the other metrics are close. With refinement, RABIT's selection rule picks
different levels $\eta$ ($\lambda$ in their paper) in our runs and in theirs (Table~\ref{tab:t9rule}). At a
matching $\eta$, our RABIT\,+\,refinement lies within $0.006$ of the published
accuracy and recourse ratio (ours vs.\ published) on FICO ($\eta=0.01$: $0.714$ /
$0.865$ vs.\ $0.716$ / $0.859$), COMPAS ($\eta=0.32$: $0.674$ / $0.987$ vs.\
$0.677$ / $0.986$) and Bail ($\eta=0.32$: $0.700$ / $0.867$ vs.\ $0.706$ /
$0.868$). On Adult, at the published accuracy ($\eta=0.08$), it gives $0.850$ /
$0.781$ against $0.850$ / $0.847$, a gap of about one standard deviation over
our ten splits ($0.060$).

\begin{table}[!htb]
\centering
\caption{Reproduction of RABIT's published results with its released code: our
runs (ten splits, seeds 300--309) against the values published by
\citet{kanamori2025rabit}. Accuracy and recourse ratio are from their Table~9;
cost, sparsity and plausibility of the found actions from their Table~1; ROC-AUC
and F1 from the results file behind their Fig.~5 in the released repository.}
\label{tab:rabit}
\small
\begin{tabular}{lllccccccc}
\toprule
Data & Method & Values & Accuracy & Recourse & Cost & Sparsity & Plausibility & ROC-AUC & F1\\
\midrule
FICO & Vanilla & ours & 0.734 & 0.542 & 0.372 & 2.24 & 0.440 & 0.801 & 0.717 \\
 &  & published & 0.735 & 0.541 & 0.358 & 2.16 & 0.438 & 0.802 & 0.717 \\
 & RABIT ($\gamma=0.002$) & ours & 0.728 & 0.832 & 0.167 & 1.05 & 0.475 & 0.796 & 0.711 \\
 &  & published & 0.732 & 0.825 & 0.165 & 1.08 & 0.475 & 0.796 & 0.711 \\
\midrule
COMPAS & Vanilla & ours & 0.684 & 0.826 & 0.232 & 1.51 & 0.436 & 0.736 & 0.727 \\
 &  & published & 0.682 & 0.832 & 0.230 & 1.52 & 0.436 & 0.734 & 0.728 \\
 & RABIT ($\gamma=0.002$) & ours & 0.683 & 0.932 & 0.108 & 1.20 & 0.471 & 0.734 & 0.726 \\
 &  & published & 0.677 & 0.936 & 0.112 & 1.26 & 0.466 & 0.733 & 0.727 \\
\midrule
Adult & Vanilla & ours & 0.853 & 0.273 & 0.353 & 1.50 & 0.476 & 0.909 & 0.632 \\
 &  & published & 0.852 & 0.274 & 0.354 & 1.47 & 0.477 & 0.907 & 0.632 \\
 & RABIT ($\gamma=0.002$) & ours & 0.852 & 0.437 & 0.281 & 1.42 & 0.478 & 0.906 & 0.633 \\
 &  & published & 0.851 & 0.427 & 0.275 & 1.42 & 0.479 & 0.905 & 0.633 \\
\midrule
Bail & Vanilla & ours & 0.707 & 0.366 & 0.456 & 1.44 & 0.507 & 0.739 & 0.499 \\
 &  & published & 0.711 & 0.364 & 0.449 & 1.55 & 0.508 & 0.735 & 0.493 \\
 & RABIT ($\gamma=0.002$) & ours & 0.702 & 0.637 & 0.225 & 1.13 & 0.517 & 0.729 & 0.474 \\
 &  & published & 0.701 & 0.628 & 0.225 & 1.13 & 0.519 & 0.725 & 0.469 \\
\bottomrule
\end{tabular}
\end{table}

\begin{table}[!htb]
\centering
\caption{Table~\ref{tab:rabitbench} with standard deviations over the ten splits
for every metric, in the format of \citet[Table~6]{kanamori2025rabit}, and
training time in seconds \citep[cf.][Fig.~2]{kanamori2025rabit}. Times are
wall-clock on shared cloud CPUs with four runs in parallel, so only ratios
between methods on the same data set are meaningful. On Bail the selection rule
picks $\gamma=0.002$, so the two RABIT rows coincide.}
\label{tab:rabitsd}
\scriptsize
\setlength{\tabcolsep}{3pt}
\begin{tabular}{llcccccccr}
\toprule
Data & Method & Accuracy & Recourse & Cost & Sparsity & Plausibility & ROC-AUC & F1 & Time (s)\\
\midrule
FICO & Vanilla & $0.734_{\pm.006}$ & $0.542_{\pm.008}$ & $0.372_{\pm.017}$ & $2.24_{\pm0.14}$ & $0.440_{\pm.004}$ & $0.801_{\pm.008}$ & $0.717_{\pm.008}$ & $96_{\pm1}$ \\
 & OAF & $0.722_{\pm.009}$ & $0.628_{\pm.010}$ & $0.316_{\pm.006}$ & $2.77_{\pm0.08}$ & $0.443_{\pm.005}$ & $0.786_{\pm.009}$ & $0.704_{\pm.011}$ & $90_{\pm12}$ \\
 & RABIT ($\gamma=0.002$) & $0.728_{\pm.009}$ & $0.832_{\pm.014}$ & $0.167_{\pm.013}$ & $1.05_{\pm0.02}$ & $0.475_{\pm.003}$ & $0.796_{\pm.008}$ & $0.711_{\pm.010}$ & $97_{\pm4}$ \\
 & RABIT (tuned $\gamma$) & $0.727_{\pm.008}$ & $0.900_{\pm.014}$ & $0.113_{\pm.007}$ & $1.02_{\pm0.01}$ & $0.475_{\pm.003}$ & $0.790_{\pm.008}$ & $0.708_{\pm.009}$ & $71_{\pm24}$ \\
 & \ours & $0.728_{\pm.010}$ & $0.992_{\pm.002}$ & $0.052_{\pm.003}$ & $1.05_{\pm0.01}$ & $0.465_{\pm.003}$ & $0.794_{\pm.009}$ & $0.717_{\pm.010}$ & $66_{\pm9}$ \\
\midrule
COMPAS & Vanilla & $0.684_{\pm.008}$ & $0.826_{\pm.009}$ & $0.232_{\pm.017}$ & $1.51_{\pm0.06}$ & $0.436_{\pm.005}$ & $0.736_{\pm.012}$ & $0.727_{\pm.009}$ & $9_{\pm1}$ \\
 & OAF & $0.685_{\pm.010}$ & $0.854_{\pm.019}$ & $0.179_{\pm.013}$ & $1.55_{\pm0.07}$ & $0.445_{\pm.004}$ & $0.734_{\pm.013}$ & $0.726_{\pm.011}$ & $9_{\pm0}$ \\
 & RABIT ($\gamma=0.002$) & $0.683_{\pm.009}$ & $0.932_{\pm.021}$ & $0.108_{\pm.013}$ & $1.20_{\pm0.06}$ & $0.471_{\pm.007}$ & $0.734_{\pm.014}$ & $0.726_{\pm.011}$ & $9_{\pm1}$ \\
 & RABIT (tuned $\gamma$) & $0.681_{\pm.008}$ & $0.966_{\pm.009}$ & $0.087_{\pm.003}$ & $1.09_{\pm0.03}$ & $0.486_{\pm.002}$ & $0.729_{\pm.015}$ & $0.726_{\pm.009}$ & $9_{\pm0}$ \\
 & \ours & $0.680_{\pm.009}$ & $0.990_{\pm.007}$ & $0.071_{\pm.003}$ & $1.08_{\pm0.02}$ & $0.491_{\pm.005}$ & $0.731_{\pm.014}$ & $0.734_{\pm.008}$ & $9_{\pm1}$ \\
\midrule
Adult & Vanilla & $0.853_{\pm.002}$ & $0.273_{\pm.010}$ & $0.353_{\pm.007}$ & $1.50_{\pm0.12}$ & $0.476_{\pm.004}$ & $0.909_{\pm.003}$ & $0.632_{\pm.007}$ & $167_{\pm12}$ \\
 & OAF & $0.833_{\pm.004}$ & $0.307_{\pm.007}$ & $0.309_{\pm.005}$ & $1.51_{\pm0.11}$ & $0.462_{\pm.004}$ & $0.858_{\pm.003}$ & $0.554_{\pm.011}$ & $173_{\pm23}$ \\
 & RABIT ($\gamma=0.002$) & $0.852_{\pm.002}$ & $0.437_{\pm.031}$ & $0.281_{\pm.012}$ & $1.42_{\pm0.10}$ & $0.478_{\pm.004}$ & $0.906_{\pm.004}$ & $0.633_{\pm.006}$ & $169_{\pm17}$ \\
 & RABIT (tuned $\gamma$) & $0.848_{\pm.003}$ & $0.538_{\pm.023}$ & $0.240_{\pm.010}$ & $1.30_{\pm0.12}$ & $0.479_{\pm.003}$ & $0.898_{\pm.004}$ & $0.627_{\pm.008}$ & $159_{\pm22}$ \\
 & \ours & $0.850_{\pm.002}$ & $0.999_{\pm.001}$ & $0.083_{\pm.036}$ & $1.35_{\pm0.13}$ & $0.439_{\pm.004}$ & $0.904_{\pm.003}$ & $0.633_{\pm.007}$ & $186_{\pm54}$ \\
\midrule
Bail & Vanilla & $0.707_{\pm.006}$ & $0.366_{\pm.020}$ & $0.456_{\pm.008}$ & $1.44_{\pm0.05}$ & $0.507_{\pm.003}$ & $0.739_{\pm.009}$ & $0.499_{\pm.012}$ & $18_{\pm1}$ \\
 & OAF & $0.678_{\pm.006}$ & $0.410_{\pm.021}$ & $0.371_{\pm.010}$ & $1.60_{\pm0.08}$ & $0.502_{\pm.004}$ & $0.706_{\pm.009}$ & $0.453_{\pm.016}$ & $18_{\pm1}$ \\
 & RABIT ($\gamma=0.002$) & $0.702_{\pm.005}$ & $0.637_{\pm.025}$ & $0.225_{\pm.026}$ & $1.13_{\pm0.04}$ & $0.517_{\pm.003}$ & $0.729_{\pm.009}$ & $0.474_{\pm.018}$ & $18_{\pm0}$ \\
 & RABIT (tuned $\gamma$) & $0.702_{\pm.005}$ & $0.637_{\pm.025}$ & $0.225_{\pm.026}$ & $1.13_{\pm0.04}$ & $0.517_{\pm.003}$ & $0.729_{\pm.009}$ & $0.474_{\pm.018}$ & $18_{\pm0}$ \\
 & \ours & $0.702_{\pm.007}$ & $0.769_{\pm.027}$ & $0.167_{\pm.008}$ & $1.07_{\pm0.03}$ & $0.513_{\pm.003}$ & $0.731_{\pm.009}$ & $0.511_{\pm.019}$ & $30_{\pm3}$ \\
\bottomrule
\end{tabular}
\end{table}

\textbf{Selection rule and refinement. }
Table~\ref{tab:tol} re-selects both methods with accuracy tolerances from
$0.5$ to $2$ points. Table~\ref{tab:refgrid} compares the refined methods at
every level $\eta$, and Table~\ref{tab:t9rule} applies RABIT's own rule for
choosing $\eta$.

\begin{table}[!htb]
\centering
\caption{Sensitivity of the shared selection rule under RABIT's protocol to its
accuracy tolerance (points below Vanilla's validation accuracy). For each
tolerance both methods are re-selected by the same rule; entries are \ours minus
tuned RABIT on test, paired over ten splits ($^*$: $p<0.05$). The only
significant accuracy loss is COMPAS at $2$ points ($-0.007$), within the
tolerance the rule allows. $^\dagger$Cost-model artifact.}
\label{tab:tol}
\small
\begin{tabular}{lcccccccc}
\toprule
& \multicolumn{2}{c}{FICO} & \multicolumn{2}{c}{COMPAS} & \multicolumn{2}{c}{Bail} & \multicolumn{2}{c}{Adult$^\dagger$}\\
Tolerance & $\Delta$acc. & $\Delta$rec. & $\Delta$acc. & $\Delta$rec. & $\Delta$acc. & $\Delta$rec. & $\Delta$acc. & $\Delta$rec.\\
\midrule
0.5 & $+0.004$ & $+0.081^*$ & $+0.002$ & $-0.015$ & $+0.000$ & $+0.235^*$ & $+0.002^*$ & $+0.461^*$ \\
0.75 & $+0.001$ & $+0.092^*$ & $+0.003$ & $+0.005$ & $-0.001$ & $+0.280^*$ & $+0.002^*$ & $+0.461^*$ \\
1 (paper) & $+0.001$ & $+0.092^*$ & $0.000$ & $+0.024^*$ & $+0.000$ & $+0.132^*$ & $+0.002^*$ & $+0.461^*$ \\
1.5 & $+0.001$ & $+0.092^*$ & $0.000$ & $+0.024^*$ & $-0.002$ & $+0.151^*$ & $+0.002^*$ & $+0.461^*$ \\
2 & $+0.001$ & $+0.092^*$ & $-0.007^*$ & $+0.034^*$ & $-0.002$ & $+0.151^*$ & $+0.002^*$ & $+0.461^*$ \\
\bottomrule
\end{tabular}
\end{table}

\begin{table}[!htb]
\centering
\caption{CERBO\,+\,refinement minus RABIT\,+\,refinement, with RABIT at its
published $\gamma=0.002$, at the same refinement level $\eta$ (RABIT's protocol; mean of the paired difference over ten
splits). $^\circ$Recourse difference with $p\ge0.05$ (paired $t$-test); all
others have $p<0.05$. No accuracy difference at $\eta>0$ is significantly
negative. $^\dagger$Cost-model artifact.}
\label{tab:refgrid}
\small
\begin{tabular}{lcccccccc}
\toprule
& \multicolumn{2}{c}{FICO} & \multicolumn{2}{c}{COMPAS} & \multicolumn{2}{c}{Bail} & \multicolumn{2}{c}{Adult$^\dagger$}\\
$\eta$ & $\Delta$acc. & $\Delta$rec. & $\Delta$acc. & $\Delta$rec. & $\Delta$acc. & $\Delta$rec. & $\Delta$acc. & $\Delta$rec.\\
\midrule
0 (none) & $+0.001$ & $+0.159$ & $-0.003$ & $+0.058$ & $+0.000$ & $+0.132$ & $-0.002$ & $+0.562$ \\
0.01 & $+0.004$ & $+0.095$ & $+0.001$ & $+0.034$ & $+0.010$ & $+0.084$ & $+0.001$ & $+0.515$ \\
0.02 & $+0.004$ & $+0.075$ & $+0.001$ & $+0.028$ & $+0.011$ & $+0.073$ & $+0.002$ & $+0.415$ \\
0.04 & $+0.004$ & $+0.061$ & $+0.001$ & $+0.021$ & $+0.012$ & $+0.069$ & $+0.004$ & $+0.314$ \\
0.08 & $+0.005$ & $+0.052$ & $+0.001$ & $+0.015$ & $+0.009$ & $+0.067$ & $+0.007$ & $+0.219$ \\
0.16 & $+0.005$ & $+0.044$ & $+0.002$ & $+0.008$ & $+0.012$ & $+0.063$ & $+0.014$ & $+0.116$ \\
0.32 & $+0.006$ & $+0.032$ & $+0.001$ & $+0.005$$^\circ$ & $+0.017$ & $+0.052$ & $+0.032$ & $+0.054$ \\
0.64 & $+0.013$ & $+0.019$ & $+0.003$ & $+0.001$$^\circ$ & $+0.033$ & $+0.029$ & $+0.050$ & $+0.029$ \\
1.28 & $+0.021$ & $+0.009$ & $+0.004$ & $+0.002$ & $+0.047$ & $+0.009$ & $+0.076$ & $+0.014$ \\
\bottomrule
\end{tabular}
\end{table}

\begin{table}[!htb]
\centering
\caption{The selection rule of \citet[App.~C.6, Table~9]{kanamori2025rabit}:
for each method, the $\eta>0$ whose test accuracy is closest to unrefined
RABIT's (accuracy, recourse ratio and the chosen $\eta$). The rule is
sensitive to near-ties: on COMPAS RABIT's accuracy is $0.677$ for every
$\eta\in[0.01,0.08]$, so it selects $\eta=0.01$ in our run and $\eta=0.32$ in the
published run. $^\dagger$Cost-model artifact.}
\label{tab:t9rule}
\small
\setlength{\tabcolsep}{4pt}
\begin{tabular}{lcccccccccccc}
\toprule
& \multicolumn{3}{c}{FICO} & \multicolumn{3}{c}{COMPAS} & \multicolumn{3}{c}{Bail} & \multicolumn{3}{c}{Adult$^\dagger$}\\
\cmidrule(lr){2-4}\cmidrule(lr){5-7}\cmidrule(lr){8-10}\cmidrule(l){11-13}
Method & Acc. & Rec. & $\eta$ & Acc. & Rec. & $\eta$ & Acc. & Rec. & $\eta$ & Acc. & Rec. & $\eta$\\
\midrule
Vanilla\,+\,ref. & 0.709 & 0.696 & 0.01 & 0.679 & 0.888 & 0.01 & 0.697 & 0.827 & 0.32 & 0.854 & 0.682 & 0.04 \\
RABIT\,+\,ref.   & 0.714 & 0.865 & 0.01 & 0.677 & 0.932 & 0.01 & 0.700 & 0.867 & 0.32 & 0.853 & 0.686 & 0.04 \\
\ours\,+\,ref.   & 0.718 & 0.959 & 0.01 & 0.678 & 0.977 & 0.04 & 0.701 & 0.959 & 0.64 & 0.857 & 1.000 & 1.28 \\
\bottomrule
\end{tabular}
\end{table}

\FloatBarrier
\subsection{Frontiers under Additive Costs}\label{app:fr}
\begin{figure}[!htb]
\centering
\includegraphics[width=\linewidth]{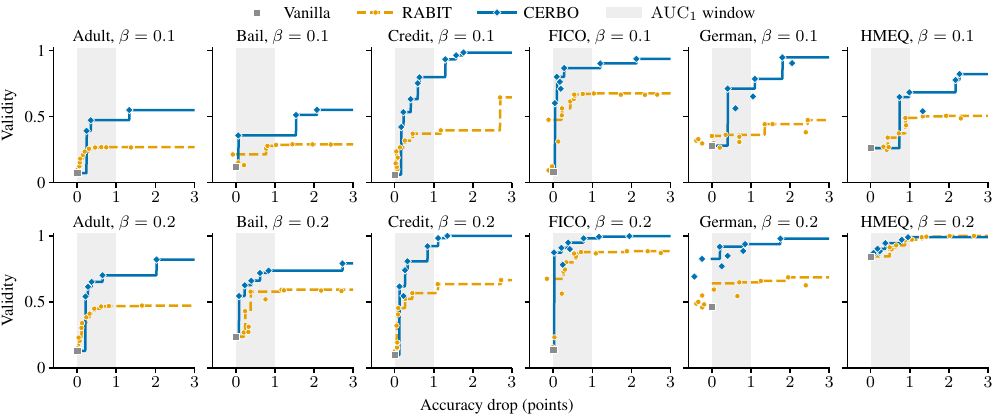}
\caption{Accuracy--recourse frontiers (validity) for all 12 settings of
Table~\ref{tab:main} (additive costs, test data, mean over seeds 200--207),
drawn as in Figure~\ref{fig:frontiers}: markers are trained models, lines are
model frontiers, and the gray band is the window of $\auc$. Models outside the
plotted range of accuracy drops are not drawn but count toward the frontiers.
German and HMEQ were added after method development. On HMEQ at $\beta=0.2$ both methods reach a
validity near 1 within the window.}
\label{fig:frontiers-all}
\end{figure}

\textbf{All frontiers. }
Figure~\ref{fig:frontiers-all} shows the frontiers behind Table~\ref{tab:main}
for all 12 settings, drawn as in Figure~\ref{fig:frontiers}.
Table~\ref{tab:window} varies the window $\Delta$ of $\auc$, and
Table~\ref{tab:refine} applies RABIT's leaf refinement to every trained model
along the trade-off.

\textbf{Decision-threshold shifts. }
RABIT's second post-processing raises the model's intercept so that more people
are approved; it is also the simplest alternative to recourse-aware training.
On fresh seeds 900--904 we evaluate every Vanilla and RABIT model at eleven
intercept shifts from $0$ to $3$, a superset of the shifts RABIT's own rule can
choose, and take the frontier over all of them. \ours without a shift still has
the larger $\auc$ in 11 of 12 settings, with the interval above zero in 9 (Credit,
$\beta=0.1$: $0.751$ vs.\ $0.402$). A shift approves everyone a little more
easily and pays for it in accuracy; recourse-aware training places favorable
leaves where affordable actions lead. Letting \ours shift as well adds at most
$0.03$.

\begin{table}[!htb]
\centering
\caption{Window sensitivity of Table~\ref{tab:main}: \ours minus RABIT in the
area under the accuracy--recourse frontier (validity) up to an accuracy drop of
$\Delta=0.5$, $1$ and $2$ points (envelope, test data, seeds 200--207, RABIT with 18 weights; 95\% paired
bootstrap intervals). The point estimate is positive in every setting at every
window; the interval is above zero in 9, 11 and 11 of the 12 settings.}
\label{tab:window}
\small
\begin{tabular}{lccc}
\toprule
Setting & $\Delta=0.5$ & $\Delta=1$ & $\Delta=2$\\
\midrule
Adult $\beta=0.1$ & $+0.124$ [$+0.105$, $+0.149$] & $+0.180$ [$+0.164$, $+0.203$] & $+0.230$ [$+0.212$, $+0.249$] \\
Adult $\beta=0.2$ & $+0.141$ [$+0.082$, $+0.216$] & $+0.191$ [$+0.160$, $+0.232$] & $+0.246$ [$+0.229$, $+0.268$] \\
Bail $\beta=0.1$ & $+0.129$ [$+0.013$, $+0.209$] & $+0.143$ [$+0.072$, $+0.207$] & $+0.179$ [$+0.131$, $+0.233$] \\
Bail $\beta=0.2$ & $+0.146$ [$+0.005$, $+0.316$] & $+0.144$ [$+0.068$, $+0.258$] & $+0.154$ [$+0.113$, $+0.217$] \\
Credit $\beta=0.1$ & $+0.176$ [$+0.117$, $+0.309$] & $+0.292$ [$+0.233$, $+0.376$] & $+0.372$ [$+0.253$, $+0.443$] \\
Credit $\beta=0.2$ & $+0.183$ [$+0.112$, $+0.277$] & $+0.244$ [$+0.208$, $+0.302$] & $+0.298$ [$+0.279$, $+0.325$] \\
FICO $\beta=0.1$ & $+0.315$ [$+0.114$, $+0.377$] & $+0.265$ [$+0.167$, $+0.302$] & $+0.252$ [$+0.200$, $+0.274$] \\
FICO $\beta=0.2$ & $+0.287$ [$-0.026$, $+0.320$] & $+0.194$ [$+0.039$, $+0.215$] & $+0.154$ [$+0.076$, $+0.166$] \\
\midrule
German $\beta=0.1$ & $+0.171$ [$+0.007$, $+0.448$] & $+0.269$ [$+0.069$, $+0.469$] & $+0.360$ [$+0.170$, $+0.490$] \\
German $\beta=0.2$ & $+0.272$ [$-0.001$, $+0.400$] & $+0.282$ [$+0.090$, $+0.391$] & $+0.289$ [$+0.174$, $+0.368$] \\
HMEQ $\beta=0.1$ & $+0.067$ [$+0.043$, $+0.141$] & $+0.121$ [$+0.090$, $+0.172$] & $+0.168$ [$+0.152$, $+0.202$] \\
HMEQ $\beta=0.2$ & $+0.033$ [$-0.003$, $+0.063$] & $+0.030$ [$+0.000$, $+0.050$] & $+0.014$ [$-0.002$, $+0.024$] \\
\bottomrule
\end{tabular}
\end{table}

\begin{table}[!htb]
\centering
\caption{RABIT's leaf refinement applied to every trained model along the
trade-off (Bail and FICO; seeds 300--309 under the max cost, 200--204 under the
additive cost). $\auc$ on validity (higher is better); each ``+\,ref.''
frontier uses all trained models and all $\eta$. $\Delta$ is \ours\,+\,ref.\
minus RABIT\,+\,ref., the stronger refined baseline in every setting, with a 95\%
paired bootstrap interval. The best $\auc$ per row is in bold.}
\label{tab:refine}
\small
\begin{tabular}{lcccl}
\toprule
Setting & RABIT\,+\,ref. & \ours & \ours\,+\,ref. & $\Delta$ \\
\midrule
Bail, max cost, $\beta=0.2$     & 0.686 & 0.607 & \textbf{0.843} & $+0.157$ [$+0.053$, $+0.292$] \\
FICO, max cost, $\beta=0.2$     & 0.568 & \textbf{0.860} & \textbf{0.860} & $+0.292$ [$+0.192$, $+0.376$] \\
Bail, additive, $\beta=0.1$     & 0.378 & 0.237 & \textbf{0.564} & $+0.186$ [$+0.114$, $+0.353$] \\
FICO, additive, $\beta=0.1$     & 0.628 & 0.834 & \textbf{0.847} & $+0.220$ [$+0.108$, $+0.336$] \\
\bottomrule
\end{tabular}
\end{table}

\FloatBarrier
\subsection{Certified and Guaranteed Rates}\label{app:cr}
\begin{figure}[!htb]
\begin{minipage}[t]{0.485\textwidth}
\vspace{0pt}\centering
\captionof{table}{Test-time certified recourse under additive costs (seeds 400--404).
Certified-rate $\auc$ (higher is better) for RABIT and \ours, their
difference with the lower end of its 95\% paired bootstrap interval, and the
mean absolute error (MAE) between \ours's requested rate $\rho$ and its test
certified rate, over requested rates whose model is within 2 points of Vanilla's
accuracy. \ours leads in 11 of 12 settings.}
\label{tab:cert}
\small
\setlength{\tabcolsep}{4pt}
\begin{tabular}{lcccc}
\toprule
Setting & RABIT & \ours & $\Delta$ [lower] & MAE $\downarrow$ \\
\midrule
Adult $0.1$  & 0.203 & \textbf{0.332} & $+0.130$ [$+0.109$] & 0.050 \\
Adult $0.2$  & 0.286 & \textbf{0.449} & $+0.164$ [$+0.113$] & 0.053 \\
Bail $0.1$   & 0.149 & \textbf{0.233} & $+0.084$ [$+0.004$] & 0.027 \\
Bail $0.2$   & 0.300 & \textbf{0.416} & $+0.116$ [$-0.016$] & 0.062 \\
Credit $0.1$ & 0.278 & \textbf{0.382} & $+0.103$ [$+0.088$] & 0.012 \\
Credit $0.2$ & 0.299 & \textbf{0.538} & $+0.239$ [$+0.187$] & 0.012 \\
FICO $0.1$   & 0.316 & \textbf{0.625} & $+0.309$ [$+0.141$] & 0.048 \\
FICO $0.2$   & 0.358 & \textbf{0.741} & $+0.383$ [$+0.216$] & 0.057 \\
German $0.1$ & 0.166 & \textbf{0.468} & $+0.302$ [$+0.209$] & 0.047 \\
German $0.2$ & 0.396 & \textbf{0.603} & $+0.206$ [$-0.030$] & 0.056 \\
HMEQ $0.1$   & \textbf{0.221} & 0.175 & $-0.046$ [$-0.149$] & 0.019 \\
HMEQ $0.2$   & 0.663 & \textbf{0.773} & $+0.110$ [$+0.035$] & 0.090 \\
\bottomrule
\end{tabular}
\end{minipage}\hfill
\begin{minipage}[t]{0.485\textwidth}
\vspace{0pt}\centering
\includegraphics[width=\linewidth]{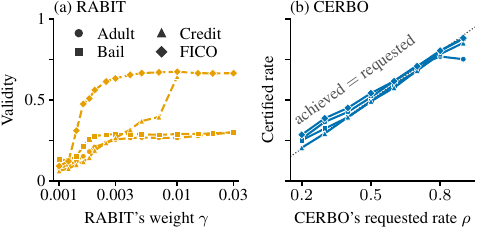}
\captionof{figure}{A weight against a requested rate on the four main data sets under
additive costs ($\beta=0.1$, test data).
(a) RABIT's validity against its weight $\gamma$ (mean over seeds 200--207) rises
over a narrow range that differs by data set and levels off near $0.3$ on Adult
and Bail and $0.67$ on FICO; missing points are models that approve everyone.
(b) \ours's certified rate on new individuals against the requested rate $\rho$
(mean over seeds 400--404) follows the dotted diagonal, except on Adult at
$\rho=0.9$, where it reaches $0.75$. Markers distinguish data sets. Weights above
$0.03$ are not shown: they add no validity on Adult and Bail and make the model
approve everyone on FICO and, up to $\gamma=0.1$, on Credit.}
\label{fig:weights}
\end{minipage}
\end{figure}

Table~\ref{tab:cert} lists the certified-rate $\auc$ behind
Section~\ref{sec:certexp}, and Figure~\ref{fig:weights} sets a weight against a
requested rate on the four main data sets.

% Table 15 and Figure 6 share one full-width float, so that they can sit beside Table 14.

\subsection{Robustness Checks}\label{app:rob}
\textbf{Base-model accuracy. }
RABIT's boosting implementation and configuration, which all methods share, is
within about one point of scikit-learn's tuned histogram gradient boosting on
five of the six data sets (Adult $0.853$ vs.\ $0.863$; Credit, FICO, German and
HMEQ within $1.2$ points). On Bail it reaches $0.700$, as RABIT reports, against
$0.851$. A stronger configuration of the same implementation (all features at
every split, 300 trees; seeds 1000--1004) raises Bail accuracy only to $0.733$,
and it widens \ours's lead to $+0.210$ [$+0.106$, $+0.355$] and $+0.370$
[$+0.227$, $+0.453$] $\auc$ at $\beta=0.1$ and $0.2$, against $+0.123$ and
$+0.265$ in RABIT's configuration. Shallower trees lower accuracy ($0.698$ at
depth 3), and at depth 4 the lead is again large ($+0.260$ and $+0.447$).

\textbf{Stricter actionability. }
Under RABIT's Adult metadata many successful actions reduce working hours, wait
five years or reduce education. With age fixed and education and working hours
increase-only (4 mutable features instead of 5; seeds 500--504), every method's validity drops, but \ours's lead
over RABIT remains: $+0.057$ [$+0.034$, $+0.079$] at $\beta=0.1$ and $+0.169$
[$+0.133$, $+0.204$] at $\beta=0.2$. On the certified rate it leads at $\beta=0.2$
($+0.089$ [$+0.068$, $+0.111$]) and ties at $\beta=0.1$.

\textbf{Commitment depth chosen on validation data. }
A variant that trains with $k=0$ but targets the depth-1 certified rate lets
validation data choose between $k=0$ and $k=1$ (seeds 700--703 on the six real
data sets, 20--22 on the 32 synthetic cells). The $k=0$ variant is rarely
preferred, as Corollary~\ref{cor:nocert} suggests: its training bound cannot see
the certificates it is asked to produce. Validation keeps $k=1$ in 11 of 12 real
settings, and the choice comes within $0.02$ of the best fixed variant in only 9
of 32 synthetic cells, so it does not remove the weakness with many mutable
features. On the same fresh seeds \ours is ahead of RABIT in all 12 real
settings, with the interval above zero in 10.

\textbf{Point masses. }
Proposition~\ref{prop:starved} predicts that exact reachability helps most over
the depth-scheduled rule when features have point masses. With the convex
surrogate this effect does not appear on the synthetic data (contrast of the
gain at 30--50\% point mass minus 0\%: $-0.011$ [$-0.039$, $+0.012$]).

% ---------------------------------------------------------------------------
\FloatBarrier
\section{Broader Impact}\label{app:impact}
Recourse-aware training aims to help people who receive an adverse automated
decision, such as a denied loan, by ensuring that an affordable change would
reverse it. \ours adds a certificate for each such person and a guaranteed rate
for the population, so an operator who promises recourse can state for whom and
how often the promise holds.

Three risks remain. A certificate is only as meaningful as the cost model and
actionability metadata behind it: if they misstate what a change costs a
person, a certified action can be unaffordable or undesirable in practice. Under
RABIT's Adult metadata, for example, many found actions reduce working hours
(Appendix~\ref{app:rob}), so deployers should review the metadata with domain
experts and the people affected. The guarantee concerns the rate over a
population; it promises recourse neither to every rejected person nor equally
across groups, and requesting rates per group, with calibration data from each
group, could address the latter. Finally, an action that changes the model's
decision need not change the outcome the model predicts, so recourse can reward
changes that game the model rather than improve the applicant's situation.

Two of our data sets, COMPAS and Bail, come from criminal justice. We use them,
as prior recourse work does, for comparability, not to endorse automated
decisions in that domain. This work introduces no new data sets or trained models.

\end{document}